\documentclass[11pt]{article}
\usepackage{xcolor}
\usepackage{multirow}
\usepackage{subfigure}
\usepackage{caption}
\usepackage{algorithm}
\usepackage{algorithmicx}
\usepackage{booktabs}
\usepackage{algpseudocode}
\usepackage{natbib}
\usepackage{subfigure}
\usepackage{caption}
\usepackage{comment}
\usepackage{amsmath,amssymb,graphicx}
\usepackage[version=3]{mhchem}
\usepackage{placeins}
\usepackage{xcolor}

\definecolor{sbblue}{rgb}{0.0, 0.0, 1.0}

\usepackage{caption}

\usepackage[normalem]{ulem}

\usepackage{url}

\def\b1e{{\mathbf e}}

\usepackage{graphicx}
\usepackage{hyperref}
\usepackage{nameref}

\date{}
\begin{document}

\title{An Additive MLP--GNN Framework for Characterizing Chemical and Structural Contributions to Aqueous Solubility}

\author{Sampreeti Bhattacharya\thanks{Current address: Pacific Northwest National Laboratory, Richland, WA, USA; email: \texttt{sampriti@alumni.unc.edu}.}
\and
Arkaprava Roy\thanks{Department of Biostatistics, University of Florida, Gainesville, Florida, USA; email: \texttt{arkaprava.roy@ufl.edu}.}}

\maketitle

\begin{abstract}
Aqueous solubility assessment is a critical step in early-stage drug discovery, but the most predictive models merge physicochemical descriptors and molecular graph information into a single representation, obscuring whether a prediction is driven by global chemistry, molecular structure, or both. We present an additive deep-learning framework with two distinct input branches that are trained jointly: physicochemical descriptors are encoded by a multilayer perceptron (the chemical branch), molecular graph topology is encoded by a graph neural network (the structural branch), and their outputs are combined through an additive model with an optional multiplicative interaction. All model parameters, including the interaction parameter when present, are optimized simultaneously using the common prediction loss. Furthermore, pretraining on the larger AqSolDB dataset and fine-tuning on the smaller BigSolDB2 dataset {lower mean test MAE and reduce run-to-run variation, supporting transfer within AqSolDB--BigSolDB2 source--target pair in this setting}. We further interpret the fitted model using best linear projections of the branch outputs, molecule-level embedding summaries across solubility classes, and atom-level GNNExplainer masks aggregated over functional groups. These analyses show that the chemical branch aligns with familiar physicochemical descriptors, while the structural branch captures graph-topological and functional-group patterns associated with solubility. Across both datasets, the framework attains competitive predictive performance while making {its branch-specific use of physicochemical descriptors and graph representations} more transparent. For the empirically preferred non-interaction model, each fitted prediction
and matched molecular-pair difference decomposes exactly into physicochemical descriptor- and
graph-branch components, providing a quantifiable account of how the model uses
the two representations. 
\end{abstract}


\section{Introduction}

Aqueous solubility is a critical physicochemical property in early-stage drug discovery, as it strongly influences oral bioavailability and formulation strategies \citep{lipinski2000drug,lipinski2002poor}. {Experimental measurements of solubility across large libraries are slow and expensive \citep{white2000high,fink2007virtual}}, motivating the widespread use of computational models for predicting solubility based on molecular structure and associated physical/chemical descriptors \citep{lovric2021machine,lusci2013deep,lee2022novel,conn2023blinded,bongini2021molecular,xiong2021graph,jiang2021could,segler2018generating,ruiz2021pharmanet,tang2020self,ramani2024graph,jorgensen2002prediction,kovdienko2010application}.

A key challenge in solubility modeling is to obtain a chemically interpretable and quantifiable representation of the molecules. Two broad {representations} are commonly used. The first consists of global physicochemical descriptors, such as partition coefficients, molecular weight, polar surface area, and hydrogen-bond counts, which summarize bulk chemical properties of the molecule. The second consists of structural information, represented by the molecular graph of atoms and bonds. Classical graph-theoretic indices, including Wiener's index \citep{wiener1947structural}, Balaban's index \citep{balaban1982distance}, Hosoya's index \citep{hosoya1971topological}, and Randi\'{c}'s index \citep{randic1975characterization}, have long been used to encode molecular topology \citep{capecchi2020one,10.1063/5.0218154}. More recently, graph neural networks (GNNs) have provided a flexible way to learn molecular representations directly from graph structure by passing information between neighboring atoms and bonds \citep{wang2022molecular,rollins2024molprop}. Many sophisticated solubility prediction models \citep{ramani2024graph,saquer2024infrared,dwivedi2023benchmarking} combine molecular information into a single input representation or a single learned embedding. Although GNN-based methods can produce accurate predictions, they do not consistently outperform descriptor-based or traditional machine-learning models on solubility tasks \citep{jiang2021could}. However, what has not kept pace with predictive performance is interpretability, as these approaches obscure whether a prediction is driven primarily by global physicochemical properties or by the molecular scaffold. Standard GNN pipelines may further entangle these {representations} by attaching chemically meaningful physicochemical descriptors, functional-group annotations, formal charges, or fragment labels directly to graph nodes and edges. Descriptor-only models have the opposite limitation: they discard explicit molecular graph structure and can be less accurate in settings where topology carries useful predictive information. Neural-network models have also been used for solubility prediction for decades, from early descriptor-based MLPs using electro-topological and topological indices~\citep{huuskonen1998aqueous,huuskonen2000estimation} to recent deep-learning models based on RDKit descriptors, SMILES, and SELFIES representations~\citep{kurotani2021solubility,ramos2024predicting}. These studies show that neural networks can achieve strong predictive accuracy, but they are hard to interpret.

This interpretability gap motivates the first contribution of this paper. We propose an additive deep-learning framework with separate descriptor and graph branches whose parameters are learned jointly. Specifically, the two {branch outputs} are combined additively, with an optional interaction term, as \(y_i = \kappa + f(G_i) + g(\mathbf{x}_i) + \tau f(G_i) g(\mathbf{x}_i) + \epsilon_i\), where \(g\) encodes the physicochemical descriptors, \(f\) encodes the molecular graph, and \(\tau\) controls the interaction between them. \citet{bhattacharya2025look} also considered this modeling formulation using radial basis function networks over fixed chemical and graph-theoretic descriptors. In this paper, we instead learn \(f(\cdot)\) and \(g(\cdot)\) end-to-end using a descriptor-based MLP and a graph neural network. By jointly estimating the two functions, we can examine the {model-assigned component from} each branch in the final prediction.

The second contribution addresses the practical challenge of solubility data with limited samples.
High-quality solubility datasets are difficult to assemble, and smaller datasets can lead to unstable training of neural network-based models. We restrict BigSolDB2 to measurements at 298--304\,K to define a
near-room-temperature subset; this restriction should not be interpreted as
an exact match to every condition represented in AqSolDB. Under this criterion, BigSolDB2~\citep{krasnov2025bigsoldb} contains a much lower proportion of molecules than AqSolDB~\citep{sorkun2019aqsoldb}.

The third contribution is a set of post hoc interpretation analyses designed to characterize what each branch has learned. a) Although the proposed architecture separates the chemical and structural branches, the outputs of the learned branch remain nonlinear summaries. We therefore use marginal best linear projections (MBLPs) inspired by \citet{semenova2021debiased} to relate the learned branch outputs to familiar physicochemical and graph-theoretic descriptors. b) We further examine molecule-level embedding summaries across solubility classes to determine whether the learned graph representation varies systematically with solubility. c) Subsequently, we apply atom-level GNNExplainer masks \citep{ying2019gnnexplainer} and aggregate them across functional groups to {highlight} structural motifs that the fitted graph branch consistently emphasizes or underemphasizes. d) Finally, a matched molecular-pair (MMP) \citep{dossetter2013matched} analysis decomposes fitted solubility differences between closely related molecules into physicochemical descriptor- and graph-branch contributions, providing molecule-level illustrations of how the two jointly trained branches contribute to the final prediction.

Together, these three components define the main goal of the paper: to develop a solubility modeling framework that remains competitive with standard molecular learning approaches while making the separate roles of physicochemical descriptors and molecular graph structure more transparent. The remainder of the paper is organized as follows. We first describe the proposed additive architecture, data sources, molecular representations, and training procedure. We then evaluate predictive performance on AqSolDB and BigSolDB2, including transfer learning from AqSolDB to BigSolDB2. Finally, we use best linear projections, molecule-level embedding summaries, and GNNExplainer-based functional-group analyses to characterize the chemical and structural information learned by the model.

The principal contribution of this work is a branch-resolved molecular
learning framework that preserves joint end-to-end training while retaining
directly inspectable descriptor- and graph-branch outputs.
The descriptor and graph branches are trained jointly end-to-end using a combined loss, but provide branch-specific outputs, $\widehat{g}(\mathbf{x})$ and $\widehat{f}(G)$, respectively. 
This differs from embedding-level fusion approaches, in which the two
representations are merged and their separate
contributions are consequently difficult to recover.

An important consequence of this formulation is that the fitted change
for a matched molecular pair $M_a\rightarrow M_b$ can be decomposed
exactly as
\(
    \Delta\widehat{y}_{a\rightarrow b}
    =
    \Delta\widehat{g}_{a\rightarrow b}
    +
    \Delta\widehat{f}_{a\rightarrow b}.
\)
The intercept and any constant offsets cancel from this comparison,
providing a direct connection between the learned representation and
chemically interpretable molecular transformations. Combined with
population-level projections onto physicochemical and graph-theoretic
descriptors and atom-level GNN explanations, this produces an
interpretation pathway spanning global molecular properties, local
structural motifs, and matched molecular changes. The resulting
components are model-assigned associations rather than uniquely
identified physical or causal effects, but they provide a direct
account of how the jointly trained model uses complementary molecular
representations. As the formulation is not specific to solubility,
it also provides a general strategy for interpreting descriptor--graph
models for other molecular-property prediction problems.

\section{Additive Deep Characterization}
We adopt an interpretable additive structure with separate descriptor and graph input branches. The separation is architectural: the full model is trained jointly end-to-end. {As the descriptors are computed from the same molecule, the two branches are not statistically independent information sources; they are distinct model pathways for two overlapping molecular representations.} In contrast to Bhattacharya et al.~\cite{bhattacharya2025look}, who use radial basis functions over fixed chemical and graph-theoretic descriptors, we parameterize the component functions with deep neural networks and train the full model end-to-end, thereby allowing us to do more in-depth analyses. Specifically, let \(\mathbf{x}_i\) denote the vector of global physicochemical descriptors for molecule \(i\) and \(G_i\) its molecular graph. The
structural branch maps \(G_i\) to a learned structural representation,
\[
u_{i1}=f(G_i),
\]
and the chemical branch maps \(\mathbf{x}_i\) to a scalar learned chemical
representation,
\[
u_{i2}=g(\mathbf{x}_i).
\]
We refer to these quantities as the learned {graph- and physicochemical descriptor-branch
outputs}, respectively. Substituting these learned representations into an interaction model gives
\begin{equation}
y_i
=
\kappa
+
g(\mathbf{x}_i)
+
f(G_i)
+
\tau g(\mathbf{x}_i)f(G_i)
+
\epsilon_i,
\qquad i=1,\ldots,n,
\label{eq:model}
\end{equation}
where \(\kappa\) is an intercept and \(\tau\) is a learnable interaction parameter. When \(\tau=0\), the fitted value is the sum, rather than the average, of the two learned branch outputs. This additive form does not mean that the branches are trained independently; both are updated jointly by minimizing the common prediction loss. When \(\tau\neq 0\), the model allows the effect of one branch to depend on the other, capturing potential synergy or attenuation between physicochemical descriptors and molecular graph structure.

\begin{figure}[htbp]
\centering
    \includegraphics[width=\textwidth]{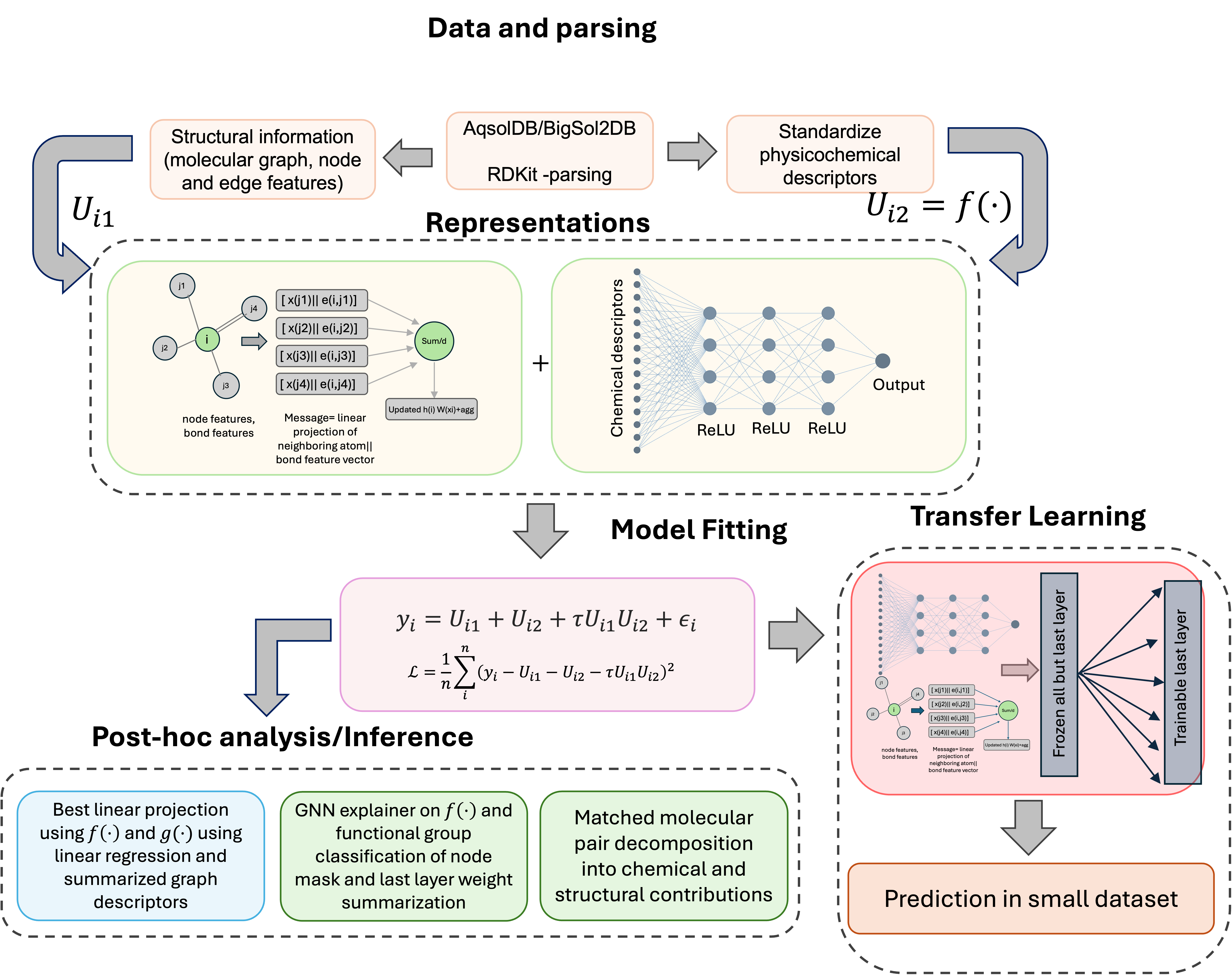}
\caption{Proposed additive deep-learning framework for aqueous solubility analysis combining two separate branches of outputs. Global physicochemical descriptors are encoded by the chemical branch via $g$, while molecular graph structure is encoded by the structural branch via $f$. The two learned branch outputs are combined through an additive model with an optional multiplicative interaction.}
\label{fig:model}
\end{figure}

In our implementation, \(g(\cdot)\) is represented by an MLP model applied to global physicochemical descriptors with details in the subsequent section. The structural component \(f(\cdot)\) is represented by a graph neural network applied to the molecular graph constructed from the SMILES string, where atoms are nodes and chemical bonds are edges. A schematic of the proposed architecture is shown in Figure~\ref{fig:model} along with analyses details. We evaluate two model configurations: an interaction model with \(\tau\neq 0\), and a no-interaction model with \(\tau=0\). Comparing these specifications assesses whether the multiplicative term adds
predictive value beyond the direct additive decomposition.

\medskip

\subsection{Data}
We analyze two aqueous-solubility datasets: AqSolDB \citep{sorkun2019aqsoldb} and BigSolDB2 \citep{krasnov2025bigsoldb}. AqSolDB contains 9,982 compounds, curated  by merging a total of nine different aqueous solubility datasets, and BigSolDB2 \citep{krasnov2025bigsoldb} contains 103{,}944 experimental
solubility values for 1{,}448 organic compounds, measured in 213 solvents over
a temperature range of 243--425\,K. As we focus on aqueous solubility, we
extract the water-solvent measurements and retain compounds in the temperature range of 
298--304\,K , yielding 692 unique SMILES after removing entries that
overlap with AqSolDB. This temperature window is chosen to match AqSolDB, whose
measurements are reported at $25\pm5\,^\circ$C. Physicochemical descriptors are
then computed with \texttt{RDKit}; a representative case is shown in SI. Each compound is represented by a SMILES string and an experimentally measured aqueous solubility value, reported as \(\log S\). The source databases aggregate measurements from different studies and do not uniformly label them as intrinsic, thermodynamic, kinetic, or measured at a common pH. We therefore treat the outcome as database-reported aqueous solubility near room temperature rather than as a homogeneous intrinsic-solubility endpoint. In particular, associations involving ionizable groups, such as carboxylic acids and amines, may partly reflect unrecorded pH, protonation-state, and assay differences and should not be interpreted as intrinsic-solubility effects. AqSolDB is used as the primary large-scale dataset for model development and pretraining, whereas BigSolDB2 is used to evaluate model performance in a smaller-sample setting and to assess transfer learning from AqSolDB.

For each compound, we compute a set of global physicochemical descriptors to define the chemical input to the MLP branch. These physicochemical descriptors include molecular weight (MolWt), octanol water partition coefficient (MolLogP), molar refractivity (MolMR), heavy-atom count (HeavyAtomCount), number of hydrogen-bond acceptors (NumHAcceptors), number of hydrogen-bond donors (NumHDonors), number of heteroatoms (NumHeteroatoms), number of rotatable bonds (NumRotatableBonds), number of valence electrons (NumValenceElectrons), number of aromatic rings (NumAromaticRings), number of saturated rings (NumSaturatedRings), number of aliphatic rings (NumAliphaticRings), total ring count (RingCount), topological polar surface area (TPSA), and Labute's approximate surface area (LabuteASA) \citep{labute2000widely}. These 15 descriptors are used as the input features for the descriptor-based physicochemical branch.

Balaban's \(J\) index~\citep{balaban1982distance} and Bertz's complexity index (BertzCT)~\citep{bertz1981first} are excluded from the MLP input because both are derived from molecular graph topology already represented by the graph branch. 

For the graph branch, each SMILES string is converted into a molecular graph, with atoms represented as nodes and chemical bonds represented as edges. Node-level features included atomic number, implicit valence, formal charge, number of radical electrons, and hybridization state. Bond information is used to define graph connectivity, with additional bond-level attributes such as bond type, conjugation, and ring membership stored as edge information. A representative example of the descriptor set and molecular graph representation is shown in Supplementary Figure~\ref{fig:expl}.
\subsection{Model architecture and computation}

The descriptor branch is implemented using \texttt{torch.nn}, and the graph branch is implemented using \texttt{torch\_geometric}. Each molecule is converted from its SMILES string into a PyTorch Geometric data object, with atoms represented as nodes and chemical bonds represented as bidirectional edges. Let \(G_i=(V_i,E_i)\) denote the molecular graph for molecule \(i\), and let \(\mathbf{x}_i\in\mathbb{R}^{p}\), with \(p=15\), denote the standardized physicochemical descriptor vector.

The graph branch takes \(G_i\) as input and produces a scalar structural output \(f(G_i)\). We consider two graph encoders: a graph convolutional network (GCN) and a message-passing neural network (MPNN). Both encoders use atom-level node features and the same graph-level pooling operation, but they differ in how node embeddings are updated within each graph layer.

For the GCN encoder, let \(\mathbf{H}^{(\ell)}\in\mathbb{R}^{N_i\times d_\ell}\) denote the matrix of node embeddings for molecule \(i\) at layer \(\ell\). The layer-wise update is
\[
\mathbf{H}^{(\ell+1)}
=
\sigma\left(
\widetilde{\mathbf{D}}^{-1/2}
\widetilde{\mathbf{A}}
\widetilde{\mathbf{D}}^{-1/2}
\mathbf{H}^{(\ell)}
\mathbf{W}^{(\ell)}
\right),
\]
where \(\widetilde{\mathbf{A}}=\mathbf{A}+\mathbf{I}\) is the adjacency matrix with self-loops, \(\widetilde{\mathbf{D}}\) is the corresponding degree matrix, \(\mathbf{W}^{(\ell)}\) is a learnable weight matrix, and \(\sigma(\cdot)\) is a nonlinear activation function. This update propagates degree-normalized information across neighboring atoms using node features and molecular connectivity, but it does not directly use bond attributes in the convolutional update.

The MPNN encoder uses an explicit bond-aware message-passing update. For each edge from atom \(u\) to atom \(v\), let \(\mathbf{e}_{uv}\) denote the corresponding edge-feature vector, containing bond-level information such as bond type, conjugation, and ring membership. Messages are formed by concatenating the source-node embedding with the edge-feature vector, applying a linear transformation, and multiplying by a degree-normalization factor:
\[
\mathbf{m}_{u\to v}^{(\ell)}
=
\alpha_{uv}
\mathbf{W}^{(\ell)}_{\mathrm{msg}}
\left[
\mathbf{h}_{u}^{(\ell)}
\,\Vert\,
\mathbf{e}_{uv}
\right],
\]
where
\[
\alpha_{uv}
=
\hat d_u^{-1/2}\hat d_v^{-1/2}
\]
is the degree-normalization factor after adding self-loops. Self-loop edges are assigned zero-valued edge features. Incoming messages are aggregated using mean aggregation, and a separate linear transformation is applied to the central node:
\[
\mathbf{h}_v^{(\ell+1)}
=
\tanh\left[
\mathbf{W}^{(\ell)}_{\mathrm{self}}\mathbf{h}_v^{(\ell)}
+
\operatorname{Mean}_{u\in\mathcal{N}(v)\cup\{v\}}
\left\{
\mathbf{m}_{u\to v}^{(\ell)}
\right\}
\right],
\]
where \(\mathcal{N}(v)\) denotes the neighbors of node \(v\). Thus, compared with the GCN encoder, the MPNN encoder makes the message construction step explicit and allows bond attributes to contribute directly to the information passed between atoms.

For both graph encoders, the graph branch consists of four graph layers. The first layer maps the node features to an embedding dimension \(d=8\), and the next three layers preserve this dimension. A hyperbolic tangent activation is applied after each graph layer. After the final graph layer, the node embeddings \(\mathbf{H}\in\mathbb{R}^{N_i\times d}\) are aggregated into a graph-level representation by concatenating global max pooling and global mean pooling:
\[
\mathbf{h}_{G_i}
=
\left[
\mathrm{GMP}(\mathbf{H})
\,\Vert\,
\mathrm{GAP}(\mathbf{H})
\right]
\in \mathbb{R}^{2d}.
\]
This pooled representation is passed through a linear output layer to produce the scalar structural branch output \(f(G_i)\). For the MPNN encoder, we also retain an atom-level local summary by summing each final node embedding across embedding dimensions; this summary is used only for downstream local-weight visualizations.

The descriptor branch takes the 15 standardized physicochemical descriptors as input. The descriptors are standardized before model fitting using \texttt{StandardScaler}. The MLP consists of three hidden layers of width \(h=4\), each followed by a ReLU activation, and a final linear output layer without bias:
\[
\begin{aligned}
\mathbf{a}^{(1)}_i &= \sigma\left(W^{(1)}\mathbf{x}_i + \mathbf{b}^{(1)}\right),\\
\mathbf{a}^{(2)}_i &= \sigma\left(W^{(2)}\mathbf{a}^{(1)}_i + \mathbf{b}^{(2)}\right),\\
\mathbf{a}^{(3)}_i &= \sigma\left(W^{(3)}\mathbf{a}^{(2)}_i + \mathbf{b}^{(3)}\right),\\
g(\mathbf{x}_i) &= W^{(4)}\mathbf{a}^{(3)}_i,
\end{aligned}
\]
where \(\sigma(t)=\max(0,t)\) is the ReLU activation. The compact hidden dimension is chosen to limit overfitting and to keep the descriptor branch from dominating the graph branch.

The final prediction combines the descriptor and graph outputs additively, with an optional multiplicative interaction:
\[
\widehat{y}_i
=
\kappa
+
g(\mathbf{x}_i)
+
f(G_i)
+
\tau g(\mathbf{x}_i)f(G_i).
\]
When the interaction model is fitted, the GNN parameters, MLP parameters, intercept, and \(\tau\) are learned simultaneously by minimizing the same loss. Thus, neither branch is fitted independently. When the no-interaction model is fitted, the multiplicative term is removed:
\[
\widehat{y}_i
=
\kappa
+
g(\mathbf{x}_i)
+
f(G_i).
\]
Comparing these two specifications allows us to assess whether the interaction term improves predictive performance beyond the simpler additive decomposition.

Although one could use the model \(\kappa + \beta_f f(G_i) + \beta_g g(x_i)\),
this does not add flexibility because $\beta_f$ and $\beta_g$ can be absorbed into the trainable output layers of $f$ and $g$. A more flexible nonlinear combination is possible, but it would make the branch contributions harder to interpret. We therefore retain the additive form and test only whether adding a multiplicative interaction improves validation and test performance.

Model parameters are estimated by minimizing the mean squared error loss,
\[
\mathcal{L}
=
\frac{1}{n}
\sum_{i=1}^{n}
\left(
y_i-\widehat{y}_i
\right)^2 .
\]
Architecture and tuning parameters, including the number of graph layers, graph embedding dimension, MLP width, and inclusion of the interaction term, are selected by minimizing validation-set prediction error. Model parameters are optimized using Adam with learning rate \(10^{-3}\). Each model is trained for 5000 epochs with a training batch size of 10; validation errors are recorded every 10 epochs using a batch size of 200. The target LogS values are kept on their original scale. To reduce overfitting, we use early stopping based on the validation loss. For each run, we select the model checkpoint corresponding to the minimum validation loss. The learned branch outputs from this checkpoint are then extracted and used for subsequent interpretability analyses. Final prediction accuracy is summarized using mean absolute error on the test set.

All experiments are repeated over 10 independent runs with different model initializations. We also examine the run-to-run correlation of the learned branch outputs as a diagnostic for the {stability of the fitted two-branch decomposition}.

\subsection{Transfer Learning from AqSolDB to BigSolDB2}

We use a lightweight transfer learning approach to improve prediction accuracy for BigSolDB2, which contains fewer samples than AqSolDB, and to assess whether molecular representations learned from the larger AqSolDB dataset {transfer to} BigSolDB2. The GNN and MLP branches pretrained on AqSolDB initialize the prediction model for BigSolDB2. During fine-tuning, all representation-learning layers remain frozen, and only the final prediction layer in each branch is updated: the graph-output layer in the GNN branch and the final fully connected layer in the MLP branch. This strategy preserves the chemical and structural representations learned from AqSolDB while allowing the final mapping to adapt to the BigSolDB2 solubility scale. Fine-tuning uses Adam and a learning rate of $10^{-3}$, with all other training settings kept the same as in the base protocol. Validation and test errors are recorded over 10 independent runs with 10 different seeds to evaluate predictive performance and run-to-run stability.

\subsection{{Marginal Best Linear Projective Analysis}}
\label{sec:mblp}

Although the proposed architecture separates chemical and structural information
into distinct branches, the learned branch outputs $f(\cdot)$ and $g(\cdot)$
remain nonlinear summaries and are therefore not directly interpretable. This is
especially true for $f(\cdot)$, which is learned from molecular graph structure
through the graph neural network. To examine what information each branch
encodes, we use marginal best linear projections (MBLPs), motivated by
\citet{semenova2021debiased}, to project the learned branch outputs onto
interpretable molecular descriptors. 

For the structural branch $f(\cdot)$, we project the learned graph-based
representation onto six graph-theoretic descriptors that are not used as direct
inputs to either branch: global efficiency, average shortest path length
(\textit{Avg\_SPL}), strength, algebraic connectivity (\textit{Alg\_connectivity}),
average degree connectivity (\textit{Avg\_degree\_connectivity}), and betweenness
centrality (\textit{Bwn\_centrality}). These projections provide a post hoc
summary of how the learned structural representation aligns with familiar
graph-level molecular summaries. For the chemical branch $g(\cdot)$, we project
the learned descriptor-based representation onto the same 15 physicochemical
descriptors used as MLP inputs. In this case, the MBLP coefficients quantify how
strongly each input descriptor is linearly reflected in the learned chemical
representation, rather than measuring encoding of unseen features.

For each of the 10 independent runs, we fit a separate univariate linear
projection, including an intercept, for each descriptor and retain only the
slope coefficient. Thus, each descriptor receives one projection coefficient per
run. As these projections are univariate and do not residualize against the
other descriptors, the resulting coefficients should be interpreted as measures
of linear association between a learned branch output and a descriptor, not as
causal effects or conditional feature importance. For the class-stratified
results in Figure~\ref{fig:mblp}, the same projections are fitted separately
within the three solubility groups defined in Section~\ref{sec:infer}. These estimates are not to be interpreted as covariate effect but $\ell_2$ projection of the estimated nonlinear function onto a linear subspace of the covariates. The full procedure is
summarized in Algorithm~\ref{algo:MBLP}.

\begin{algorithm}[htbp]
\caption{Linear Projection of Descriptors onto Learned Representations}
\label{algo:MBLP}
\begin{algorithmic}[1]
\Require Learned representations $f^{(i)}(\cdot)$ and $g^{(i)}(\cdot)$ from runs $i=1,\ldots,10$; standardized graph-theoretic descriptor matrix $\mathbf{D}_f \in \mathbb{R}^{N \times 6}$; standardized physicochemical descriptor matrix $\mathbf{D}_g \in \mathbb{R}^{N \times 15}$
\Ensure Projection coefficient matrices $C_f \in \mathbb{R}^{10 \times 6}$ and $C_g \in \mathbb{R}^{10 \times 15}$
\For{$k = 1$ \textbf{to} $10$}
    \State Load the saved representations $f^{(k)}$ and $g^{(k)}$
    \State Build the output vectors $\mathbf{f}^{(k)}$ and $\mathbf{g}^{(k)}$ over the full set of data points
    \For{$j = 1$ \textbf{to} $6$}
        \State Let $\mathbf{d}_{f,j}$ be the $j$-th column of $\mathbf{D}_f$
        \State Compute $(\widehat{\alpha}_{f,j}^{(k)},\,\widehat{\beta}_{f,j}^{(k)}) \leftarrow \arg\min_{\alpha,\beta} \|\mathbf{f}^{(k)} - \alpha\mathbf{1} - \beta \mathbf{d}_{f,j}\|_2^2$
        \State Set $C_f[k,j] \leftarrow \widehat{\beta}_{f,j}^{(k)}$
    \EndFor
    \For{$j = 1$ \textbf{to} $15$}
        \State Let $\mathbf{d}_{g,j}$ be the $j$-th column of $\mathbf{D}_g$
        \State Compute $(\widehat{\alpha}_{g,j}^{(k)},\,\widehat{\beta}_{g,j}^{(k)}) \leftarrow \arg\min_{\alpha,\beta} \|\mathbf{g}^{(k)} - \alpha\mathbf{1} - \beta \mathbf{d}_{g,j}\|_2^2$
        \State Set $C_g[k,j] \leftarrow \widehat{\beta}_{g,j}^{(k)}$
    \EndFor
\EndFor
\State \Return $C_f$, $C_g$
\end{algorithmic}
\end{algorithm}

Equivalently, for each run $(k)$, the univariate projection coefficients for the
structural descriptors are obtained from
\[
(\widehat{\alpha}_{f,j}^{(k)},\,\widehat{\beta}_{f,j}^{(k)})
=
\arg\min_{\alpha,\beta}
\left\|
\mathbf{f}^{(k)} - \alpha\mathbf{1} - \beta \mathbf{d}_{f,j}
\right\|_2^2,
\qquad
j=1,\ldots,6,
\]
and the corresponding coefficients for the physicochemical descriptors from
\[
(\widehat{\alpha}_{g,j}^{(k)},\,\widehat{\beta}_{g,j}^{(k)})
=
\arg\min_{\alpha,\beta}
\left\|
\mathbf{g}^{(k)} - \alpha\mathbf{1} - \beta \mathbf{d}_{g,j}
\right\|_2^2,
\qquad
j=1,\ldots,15.
\]
We retain only the slope coefficients $\widehat{\beta}_{f,j}^{(k)}$ and
$\widehat{\beta}_{g,j}^{(k)}$. With the intercept included, these reduce to the
mean-centered estimates
\[
\widehat{\beta}_{f,j}^{(k)}
=
\frac{(\mathbf{d}_{f,j}-\bar{d}_{f,j}\mathbf{1})^{\top}(\mathbf{f}^{(k)}-\bar{f}^{(k)}\mathbf{1})}
{(\mathbf{d}_{f,j}-\bar{d}_{f,j}\mathbf{1})^{\top}(\mathbf{d}_{f,j}-\bar{d}_{f,j}\mathbf{1})},
\qquad
\widehat{\beta}_{g,j}^{(k)}
=
\frac{(\mathbf{d}_{g,j}-\bar{d}_{g,j}\mathbf{1})^{\top}(\mathbf{g}^{(k)}-\bar{g}^{(k)}\mathbf{1})}
{(\mathbf{d}_{g,j}-\bar{d}_{g,j}\mathbf{1})^{\top}(\mathbf{d}_{g,j}-\bar{d}_{g,j}\mathbf{1})},
\]
where $\bar{d}_{\cdot,j}$, $\bar{f}^{(k)}$, and $\bar{g}^{(k)}$ denote the
corresponding sample means.

The coefficients are then collected across the 10 independent runs to form
\[
C_f =
\left(
\widehat{\beta}_{f,j}^{(k)}
\right)_{k=1,\ldots,10;\,j=1,\ldots,6},
\qquad
C_g =
\left(
\widehat{\beta}_{g,j}^{(k)}
\right)_{k=1,\ldots,10;\,j=1,\ldots,15}
\]

\subsection{Matched molecular-pair decomposition}
\label{sec:mmp_method}

To provide molecule-level interpretations of the jointly learned descriptor and
graph contributions, we perform a matched molecular-pair (MMP) analysis using
the fitted MPNN+MLP model without the interaction term. Molecular pairs are
identified using \texttt{mmpdb}. The variable fragment is restricted to at most
eight heavy atoms and to no more than 33\% of either molecule. Thus, the
conserved portion represented at least approximately two thirds of each
molecule, excluding transformations that replaced most of the molecular
structure. Pair--transformation records are enumerated using a rule-environment
radius of zero.

For model run \(k\), the fitted value for molecule \(M_i\) is
\[
\widehat{y}^{(k)}_i
=
\kappa^{(k)}
+
g^{(k)}(\mathbf{x}_i)
+
f^{(k)}(G_i),
\]
where \(g^{(k)}(\mathbf{x}_i)\) is the descriptor-branch output and
\(f^{(k)}(G_i)\) is the graph-branch output. Consequently, for a directed
matched pair \(M_a\rightarrow M_b\),
\begin{equation}
\begin{split}
\Delta\widehat{y}^{(k)}_{a\rightarrow b}
&=
\widehat{y}^{(k)}_b-\widehat{y}^{(k)}_a \\
&=
\left\{
g^{(k)}(\mathbf{x}_b)-g^{(k)}(\mathbf{x}_a)
\right\}
+
\left\{
f^{(k)}(G_b)-f^{(k)}(G_a)
\right\} \\
&=
\Delta g^{(k)}_{a\rightarrow b}
+
\Delta f^{(k)}_{a\rightarrow b}.
\end{split}
\label{eq:mmp_decomposition}
\end{equation}
The intercept cancels from the pairwise difference. The descriptor, graph, and
total pairwise changes are calculated separately for each of the 10 fitted
models and then averaged across runs. Run-to-run variability in the total
change is calculated directly from the 10 values
\(\Delta g^{(k)}+\Delta f^{(k)}\), thereby retaining the covariance between the
two jointly trained branches.

For our classification, a branch is considered approximately inactive
when the absolute mean change is below 0.15 log-solubility units and active
when it exceeds 0.40 units. A pair is classified as graph-dominated when the
descriptor branch is inactive and the graph branch is active, and as
descriptor-dominated under the converse condition. Pairs for which both
branches are active are classified as having contributions from both
branches. These categories describe the fitted additive components and do not
represent an explicit interaction between independently fitted models.

For the qualitative isomer analysis, we consider graph-dominated pairs in
which both molecules contained at least six heavy atoms, have the same molecular
formula but different canonical SMILES, and show an absolute difference of at
most 1.0 between the fitted and measured pairwise changes. Duplicate unordered
pairs were removed using their canonical SMILES.

\section{Results}
\label{sec:results}

We evaluate the proposed additive deep-learning framework from three complementary perspectives. First, we assess predictive performance on AqSolDB and BigSolDB2 across GCN and MPNN implementations of the structural branch, with and without the interaction term. Second, we examine whether representations learned from the larger AqSolDB dataset could be transferred to the substantially smaller BigSolDB2 dataset. Third, we use best linear projections, molecule-level summed embedding summaries, atom-level GNNExplainer masks, and matched-molecule pair-based analyses to characterize the chemical and structural information learned by the fitted model. 
\subsection{Predictive performance}

To assess generalizability to structurally different molecules, we employ a scaffold-splitting procedure that partitions molecules according to their two-dimensional core structural frameworks. Molecules sharing the same scaffold are assigned to the same subset, so the validation and test sets contain molecular frameworks not seen during training. This provides a more realistic estimate of model performance in prospective experimental settings, where predictions are often required for new chemical scaffolds. For AqSolDB, we borrow the scaffold split provided by \citet{dwivedi2023benchmarking} to be able to compare with their results. For BigSolDB2, we partition the data by Bemis--Murcko scaffold \citep{bemis1996properties}, computed using RDKit's \texttt{MurckoScaffold} module, following the splitting protocol of \citet{ramsundar2018molecular,wu2018moleculenet} with an 80:10:10 split for training, validation, and testing, respectively.

Table~\ref{tab:prediction} reports the mean, standard deviation, and median test MAE across 10 independent runs.
For AqSolDB, the descriptor-only MLP achieves a mean test MAE of 1.17 with standard deviation 0.08. The additive models without the interaction term consistently perform better than their corresponding interaction models. The MPNN+MLP model without interaction achieves the lowest test error, with mean MAE 1.06 and standard deviation 0.02, followed closely by the GCN+MLP model, with mean MAE 1.07 and standard deviation 0.03. Relative to the descriptor-only MLP, these models reduce mean MAE by 0.11 and 0.10, respectively, while also reducing run-to-run variability. This within-study comparison indicates that the joint model had lower mean test MAE than the 15 global physicochemical descriptors-only model. Adding the multiplicative interaction term increases the test error for both graph architectures: the GCN+MLP interaction model has mean MAE 1.12, and the MPNN+MLP interaction model has mean MAE 1.10. The interaction term therefore does not improve out-of-sample accuracy and instead adds model flexibility that is not supported by the prediction results. A plausible explanation is that the two branches encode overlapping molecular information. Their product therefore adds a redundant, higher-variance term and makes the effective gradient of each branch depend on the scale of the other branch's output. Large branch outputs can also produce disproportionately large products. These mechanisms can increase optimization instability and overfitting without adding new signal. The present results support this explanation empirically through the consistent deterioration in test MAE, although they do not establish a unique causal reason for that deterioration.

It is also important to note that \citet{dwivedi2023benchmarking} reported a GCN test MAE of approximately 1.35 on AqSolDB, whereas the present GCN+MLP model achieves a mean test MAE of 1.07. Although the comparison is not exact because of differences in implementation and repeated-run design, the improvement suggests that global physicochemical descriptors contain predictive information that is not fully captured by graph topology alone \cite{huuskonen2000estimation,ramos2024predicting}. The comparison with the descriptor-only MLP provides complementary evidence in the other direction: adding a learned graph representation improves prediction beyond the global physicochemical descriptors alone.

For BigSolDB2, the descriptor-only MLP and standalone GCN have the same mean MAE of 1.29, although the MLP is more variable across runs (standard deviation 0.18 versus 0.05). Combining the two representations improves performance: GCN+MLP reduces mean MAE to 1.24, while MPNN+MLP achieves the lowest mean MAE of 1.21. Thus, neither input representation dominates on its own, and the improvement of the joint models is consistent with a predictive benefit from combining the two representations, the physicochemical descriptor and graph branches. The larger standard deviation of 0.19 for MPNN+MLP nevertheless indicates that fitting the more flexible message-passing representation from scratch is less stable on this smaller dataset. The interaction models perform worse, with mean MAE 1.39 for both graph architectures, again indicating that the multiplicative term is not supported by these data.

\begin{table}[htbp]
\centering
\caption{Mean absolute error (MAE) performance of different descriptor and graph neural-network models on the AqSolDB and BigSolDB2 datasets. For each model configuration, results are reported as the mean MAE, standard deviation (in parentheses), and median MAE obtained over 10 independent training runs using 10 different random seeds. Lower MAE values indicate better predictive accuracy. The evaluated architectures include a descriptor-only multilayer perceptron (MLP), a baseline Graph Convolutional Network (GCN), GCN combined with a multilayer perceptron (GCN+MLP), GCN+MLP with an interaction module, Message Passing Neural Network combined with a multilayer perceptron (MPNN+MLP), and MPNN+MLP with an interaction module. The standard deviation quantifies performance variability across runs, while the median provides a robust measure of central tendency.}
\begin{tabular}{llcc}
\toprule
Dataset & Model & Mean (Std dev) & Median \\
\midrule
AqSolDB & MLP & 1.17 (0.08) & 1.15 \\
AqSolDB & GCN+MLP & 1.07 (0.03) & 1.08 \\
AqSolDB & GCN+MLP with interaction & 1.12 (0.03) & 1.12 \\
AqSolDB & MPNN+MLP & 1.06 (0.02) & 1.07 \\
AqSolDB & MPNN+MLP with interaction & 1.10 (0.05) & 1.08 \\
\midrule
BigSolDB2 & MLP & 1.29 (0.18) & 1.30 \\
BigSolDB2 & GCN & 1.29 (0.05) & 1.28 \\
BigSolDB2 & GCN+MLP & 1.24 (0.12) & 1.27 \\
BigSolDB2 & GCN+MLP with interaction & 1.39 (0.12) & 1.41 \\
BigSolDB2 & MPNN+MLP & 1.21 (0.19) & 1.21 \\
BigSolDB2 & MPNN+MLP with interaction & 1.39 (0.21) & 1.38 \\
\bottomrule
\end{tabular}
\label{tab:prediction}
\end{table}

Although the proposed models are not designed solely to optimize benchmark accuracy, their prediction errors are competitive with established GNN baselines reported by \citet{dwivedi2023benchmarking}. More importantly, the additive formulation provides a direct way to examine the model-assigned components associated with physicochemical descriptors and molecular graph structure. Thus, even when the gain in raw prediction accuracy is modest, the model offers additional interpretability by allowing the chemical and structural components to be examined separately. This is the focus of the inference analyses in Section~\ref{sec:infer}.

Overall, the \textit{best-performing model} for AqSolDB in Table~\ref{tab:prediction} is the MPNN+MLP model without the interaction term. We therefore use this model for the downstream analyses in the subsequent sections, which we base on AqSolDB rather than BigSolDB2 given the former's larger sample size and correspondingly more reliable estimates.

\subsection{Transfer learning improves prediction on BigSolDB2}

Table~\ref{tab:tl_prediction} summarizes the transfer-learning results on
BigSolDB2. Pretraining on AqSolDB and then fine-tuning on the same BigSolDB2
training, validation, and test splits used in the previous section
lowers mean test MAE relative to training directly on BigSolDB2.
After canonicalizing SMILES and computing Bemis--Murcko scaffolds, we
confirmed that the AqSolDB pretraining set and the BigSolDB2 test set are
disjoint at both the compound and scaffold levels. The best transfer-learning performance is obtained by the GCN+MLP model, which achieves mean MAE $1.03$ with standard deviation $0.07$. The MPNN+MLP transfer-learning model performed similarly, with mean MAE $1.04$ and standard deviation $0.06$. These errors are markedly lower than the corresponding models trained from scratch on BigSolDB2.

\begin{table}[htbp]
\centering
\caption{Transfer-learning performance on BigSolDB2 after pretraining on AqSolDB. For each model, the graph and descriptor branches were initialized from the AqSolDB-pretrained model and fine-tuned on BigSolDB2 by updating the final prediction layers. Values report mean, standard deviation, and median test MAE across 10 independent runs with 10 different seeds.}
\begin{tabular}{lcc}
\toprule
Model & Mean (Std dev) & Median \\
\midrule
GCN+MLP & 1.03 (0.07) & 1.03 \\
GCN+MLP with interaction & 1.09 (0.08) & 1.06 \\
MPNN+MLP & 1.04 (0.06) & 1.05 \\
MPNN+MLP with interaction & 1.11 (0.06) & 1.13 \\
\bottomrule
\end{tabular}
\label{tab:tl_prediction}
\end{table}

The improvement is especially clear for the MPNN-based model. When trained directly on BigSolDB2, MPNN+MLP had mean MAE $1.21$; with transfer learning, the mean MAE decreases to $1.04$. This provides evidence of cross-dataset transfer for this AqSolDB--BigSolDB2 comparison, where only the final prediction layers are fine-tuned. Transfer learning also reduces the variability of the MPNN-based model across independent runs, suggesting that pretraining stabilizes the learned representation and reduces sensitivity to random initialization in the small-sample setting.

Similar to the direct-training experiments, the interaction term does not improve prediction after transfer learning. The GCN+MLP transfer-learning model without interaction achieved mean MAE $1.03$, compared with $1.09$ for the corresponding interaction model. Similarly, the MPNN+MLP transfer-learning model without interaction achieved mean MAE $1.04$, compared with $1.11$  with interaction. 

\noindent
\textit{Verifying that pretraining drives the transfer gain.} To confirm that the improvement from transfer learning reflects useful representations learned on AqSolDB rather than the regularizing effect of training only a small number of parameters, we repeat the fine-tuning procedure with randomly initialized GCN, MLP, and MPNN modules: the representation layers are left untrained and frozen, and only the final prediction layer of each branch is fit on BigSolDB2. This control yields substantially higher test MAE, 1.33 (0.3) without the interaction term and 1.36 (0.3) with it, well above the 1.03--1.04 obtained when the frozen backbone is pretrained on AqSolDB. The gap indicates that the transfer-learning gain is attributable to the representations learned during pretraining, not merely to the reduced number of trainable parameters during fine-tuning.

\subsection{Post hoc interpretation analyses}
\label{sec:infer}

Aqueous solubility (\(\log S\), the base-10 logarithm of molar solubility) is commonly represented using discrete solubility classes rather than as a continuous variable. Following the classification scheme adopted in \cite{sorkun2019aqsoldb}, compounds are categorized as highly soluble (\(\log S \geq 0\)), soluble (\(-2 \leq \log S < 0\)), slightly soluble (\(-4 \leq \log S < -2\)), and insoluble (\(\log S < -4\))~\cite{sorkun2019aqsoldb}. These categories are broadly consistent with pharmaceutical development guidelines, where candidate compounds are generally required to exhibit solubilities above approximately \(10~\mu\mathrm{M}\) to support preclinical evaluation~\cite{sun2019predictive,fink2020solubility}. To identify molecular features associated with solubility extremes, we divide the dataset into three classes: poorly soluble (\(\log S \leq -5.0\)), moderately soluble (\(-5.0 < \log S < -0.5\)), and highly soluble (\(\log S \geq -0.5\)). These operational thresholds differ from the conventional boundaries of \(\log S=-4\) and \(\log S=0\). Compounds with \(\log S \leq -5.0\) fall well below the commonly accepted insolubility threshold and approach the regime of extremely low solubility~\cite{sun2019predictive}, whereas compounds with \(\log S \geq -0.5\) occupy the upper portion of the soluble range or above. As a result, compounds that would traditionally be classified as `slightly soluble'' (e.g., \(\log S \approx -2.5\)) are intentionally included within the moderate-solubility category. This broader separation of the extreme classes minimizes the inclusion of borderline compounds whose classifications may be influenced by experimental uncertainty in aggregated solubility datasets. 

\noindent
\textit{Stability of the branch-resolved decomposition.} As both branches are
fitted jointly against a single response, the additive split of predictive
signal between $f$ and $g$ is not guaranteed to be identifiable: signal shared
by the descriptor and graph information could in principle be absorbed by either
branch. {Although the descriptor and graph representations are derived from the same
molecule, they encode both overlapping and distinct molecular features.
Consequently, predictive signal accessible to both pathways may, in principle,
be allocated to either branch without changing their combined output.} We therefore check empirically how consistently each branch is recovered
across 10 independent runs (10 random seeds). For each of the
$\binom{10}{2}=45$ pairs of runs, we compute the Pearson correlation between the
per-molecule {graph-branch outputs} $f(G_i)$, and separately between the
{descriptor-branch outputs} $g(\mathbf{x}_i)$. Since the additive representation admits an offset ambiguity and the two branches can exchange overlapping predictive signal, we use correlation only as a scale-free diagnostic of reproducibility across runs, not as evidence of identifiability. As a complementary check on the relative scale of the two outputs, we compute the {structural relative-scale index}
$\rho = \mathrm{std}(f)/\bigl(\mathrm{std}(f)+\mathrm{std}(g)\bigr)$ per run,
where $\mathrm{std}(f)$ and $\mathrm{std}(g)$ are the standard deviations of the
per-molecule outputs $\{f(G_i)\}_{i=1}^{n}$ and $\{g(\mathbf{x}_i)\}_{i=1}^{n}$.
High per-branch correlations together with a stable $\rho$ provide
empirical diagnostics of reproducibility under the fitted architecture.
These diagnostics assess stability, not formal identifiability or physical separation of the two contributions.

The chemical branch $g(\mathbf{x}_i)$ is highly
reproducible (mean $r \approx 0.97$), as expected for a compact network applied
to a fixed set of 15 physicochemical descriptors. The structural branch $f(G_i)$
is correlated at $r \approx 0.80$, a moderate but expected reduction given that the graph encoder is more expressive than the descriptor branch and different initializations can settle on different but near-equivalent representations. {The relative scale of the branch outputs is stable across the fitted runs.} We obtain
$\rho = 0.33 \pm 0.03$ over the 10 runs. Here, $\mathrm{std}(f)$ and $\mathrm{std}(g)$ denote the standard deviations of the per-molecule outputs $\{f(G_i)\}_{i=1}^{n}$ and $\{g(\mathbf{x}_i)\}_{i=1}^{n}$, respectively for a given run. These results indicate that, although the fine detail of the graph
representation exhibits some run-to-run variability, the model-assigned decomposition is reasonably reproducible across the fitted runs. We therefore use it for the subsequent interpretive analyses while avoiding the stronger claim that the two components are uniquely identified chemical and structural effects.

\subsubsection{Best linear projections}
\begin{figure}
    \centering
    \includegraphics[width = 0.8\textwidth, trim=0cm 1.1cm 0cm 0cm, clip=true]{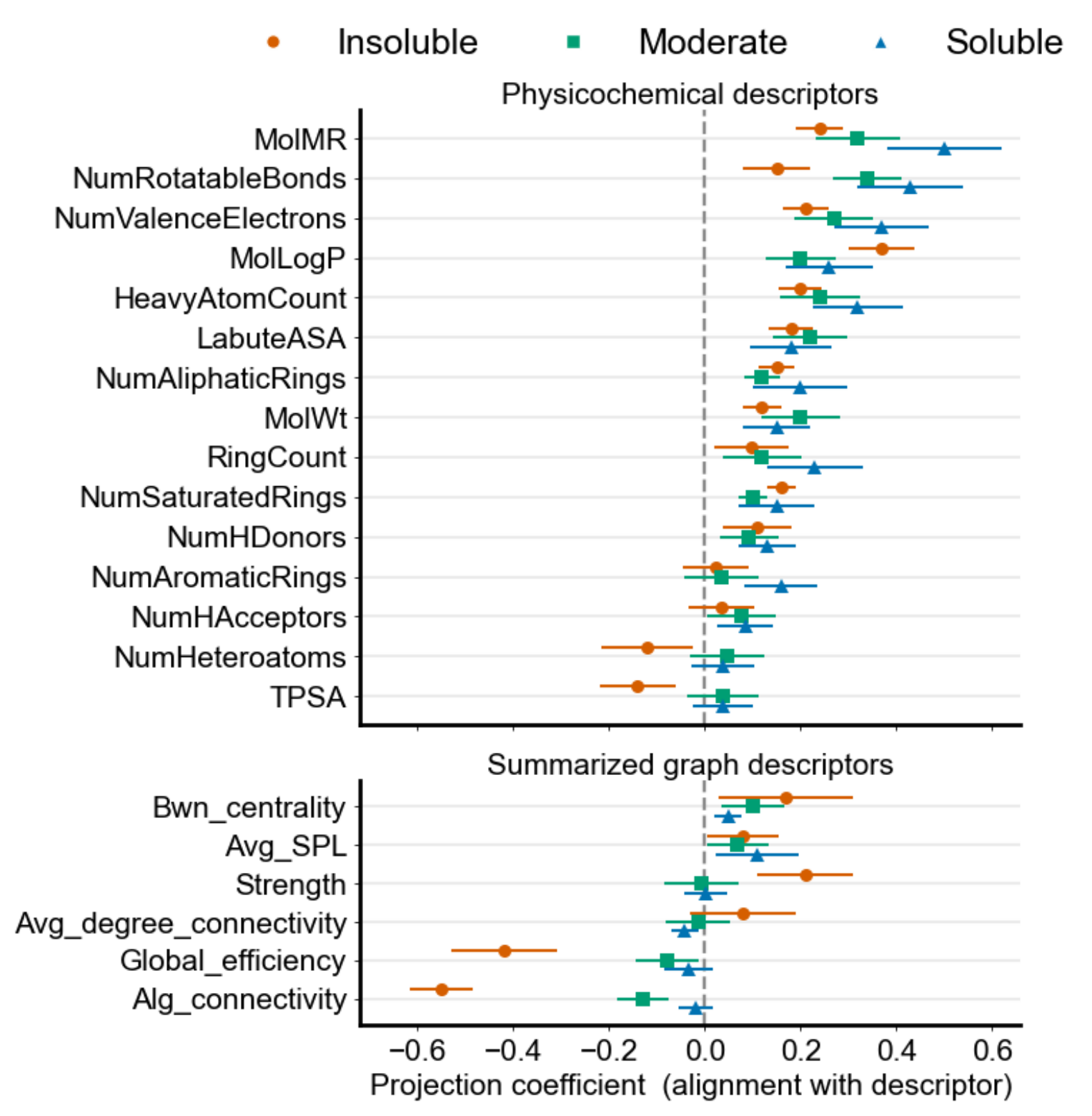}
    \caption{MBLP coefficients (calculated using the MPNN+MLP model) for the chemical (top) and structural (bottom) branches, stratified by aqueous solubility class. Points are averages over the 10 runs and bars are $\pm 1$ SD; positive values indicate that the learned summary increases with the descriptor. Associations are post hoc summaries of the learned representations, not causal effects. {The across-run SD reflects training variability rather than sampling uncertainty.}}
    \label{fig:mblp}
\end{figure}

Following the protocol in Section~\ref{sec:mblp}, we examine MBLP coefficients as univariate associations between learned branch outputs and interpretable molecular summaries. The top panel in Figure~\ref{fig:mblp} summarizes the chemical-branch MBLP coefficients stratified by solubility class. Several descriptors show stable positive associations with the learned chemical representation across solubility groups. MolMR, HeavyAtomCount, NumRotatableBonds, and NumValenceElectrons are particularly prominent, with larger coefficients among soluble compounds than among insoluble compounds. For example, MolMR increases from $0.24$ in the insoluble group to $0.50$ in the soluble group, while NumRotatableBonds increases from $0.15$ to $0.43$. These patterns indicate that the chemical branch is not simply reproducing a single descriptor, but is instead learning a composite representation that reflects molecular size, refractivity, flexibility, and electronic content. As the solubility values lie within different baseline ranges across the three outcome-defined classes, the intercept accounts for location shifts.

Several of these physicochemical descriptors, such as MolWt, MolMR, HeavyAtomCount, NumValenceElectrons, and LabuteASA, are mutually correlated through their shared dependence on molecular size, so their individually large projection coefficients should be read as a single coherent size-and-bulk signal rather than five independent contributions. That the chemical branch consistently aligns with this entire size-related cluster, positively across all independent runs and across solubility classes, is worth noting: the MLP branch reproducibly encodes molecular size and bulk as a dominant axis of its learned chemical representation.

MolLogP shows a different pattern, with the largest coefficient among insoluble compounds. This is consistent with the known role of hydrophobicity in reducing aqueous solubility, but here it should be interpreted more cautiously as a projection of the learned chemical representation rather than a direct regression effect on solubility. TPSA and NumHeteroatoms have negative projection coefficients in the insoluble group but small positive values in the moderate and soluble groups, suggesting that the chemical branch encodes these descriptors differently across three solubility ranges.

The bottom panel in Figure~\ref{fig:mblp} summarizes MBLP coefficients for graph-topology descriptors. The structural branch shows noticeable differences between insoluble and soluble compounds. Global efficiency and algebraic connectivity are strongly negative among insoluble molecules, with coefficients $-0.42$ and $-0.55$, respectively, but are much closer to zero among soluble molecules. Strength and betweenness centrality are positive in the insoluble group and smaller in the moderate and soluble groups. Since these are global graph descriptors, their signs should not be read as direct mechanistic explanations. Instead, they mainly show that the learned structural representation captures recognizable network-level properties.

 In summary, the MBLP results support the interpretability goal of the proposed architecture. The chemical branch aligns with familiar physicochemical descriptors, while the structural branch aligns with graph-topology summaries. Importantly, these associations are obtained after model fitting and are therefore summaries of the information encoded by the learned representations, not inputs artificially imposed on the final prediction.

\subsubsection{Molecule-level embedding summaries differ across solubility classes}

We next examine molecule-level {summed embedding summaries} from the fitted
MPNN+MLP model. For each molecule, this quantity is the sum over atoms of the last-layer node
embeddings, $\sum_{v \in V_i}\sum_{k=1}^{d} h^{(L)}_{v,k}$, computed after
training, where $v$ runs over the atoms of molecule $i$ and $k$ runs over the
$d$ embedding dimensions. The {summed summary} is a simple unweighted aggregate of the node embeddings,
distinct from the trained structural output $f(G_i)$. 
As it is a simple unweighted
aggregate of the node embeddings rather than the model's prediction, separation
across solubility classes is not guaranteed; on the other hand, such separation would indicate that the
atom-level summed embeddings we later interpret carry solubility-sensitive structural information.

\begin{figure*}[htbp]
\centering
\begin{minipage}{0.47\textwidth}
\centering
\includegraphics[width=\linewidth]{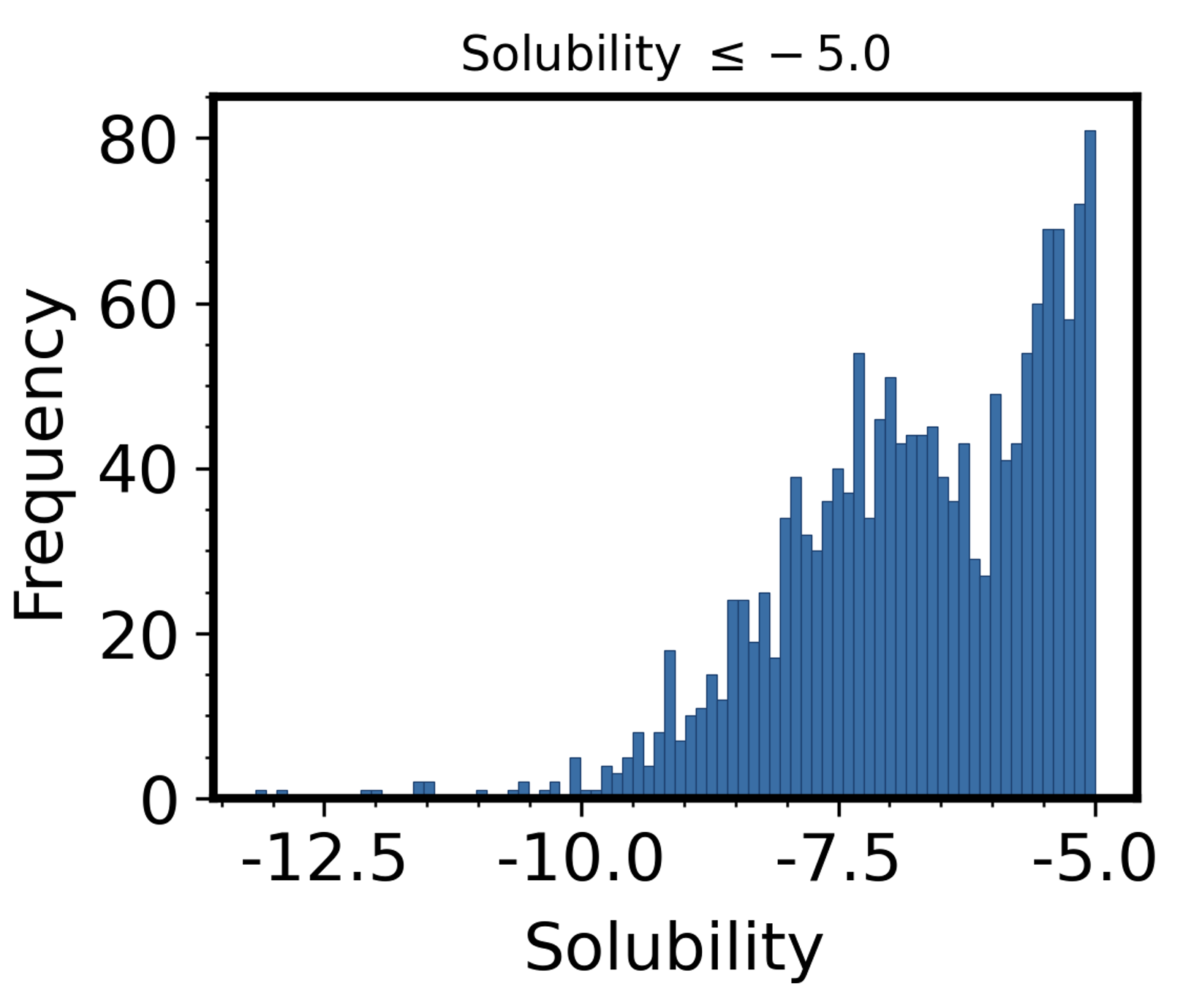}\\[-2pt]
\textnormal{(a)}
\end{minipage}
\hfill
\begin{minipage}{0.47\textwidth}
\centering
\includegraphics[width=\linewidth]{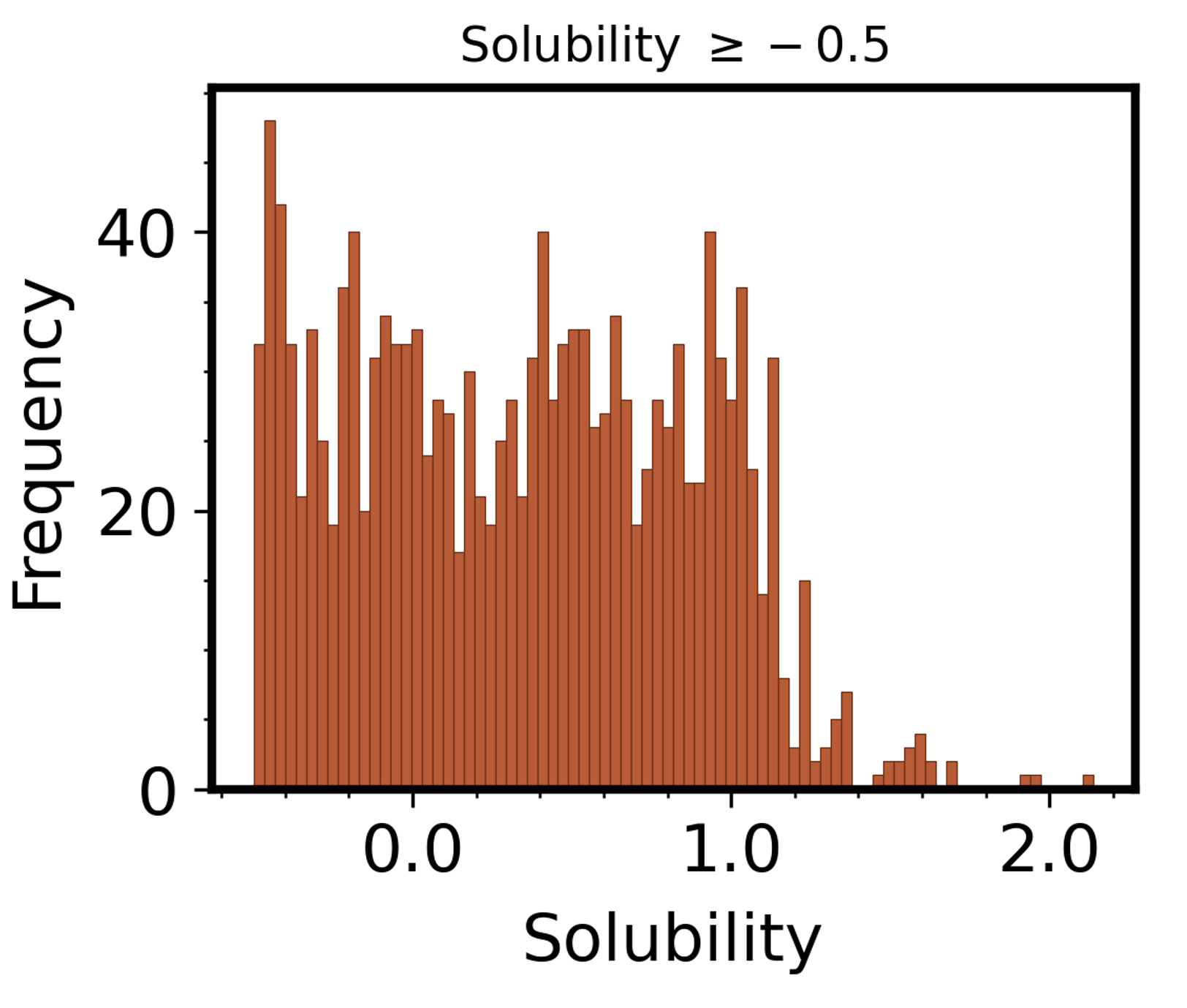}\\[-2pt]
\textnormal{(b)}
\end{minipage}

\begin{minipage}{0.47\textwidth}
\centering
\includegraphics[width=\linewidth]{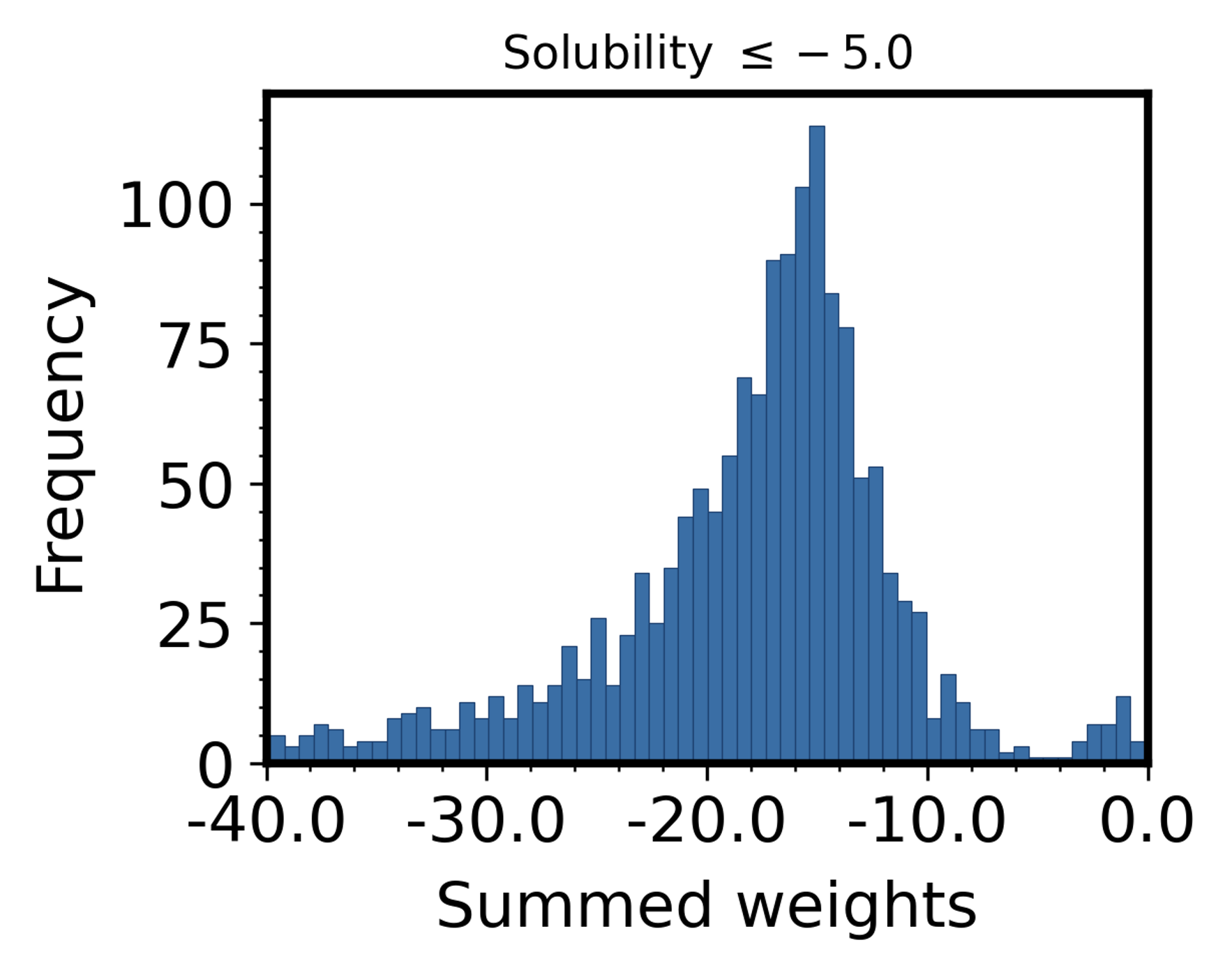}\\[-2pt]
\textnormal{(c)}
\end{minipage}
\hfill
\begin{minipage}{0.47\textwidth}
\centering
\includegraphics[width=\linewidth]{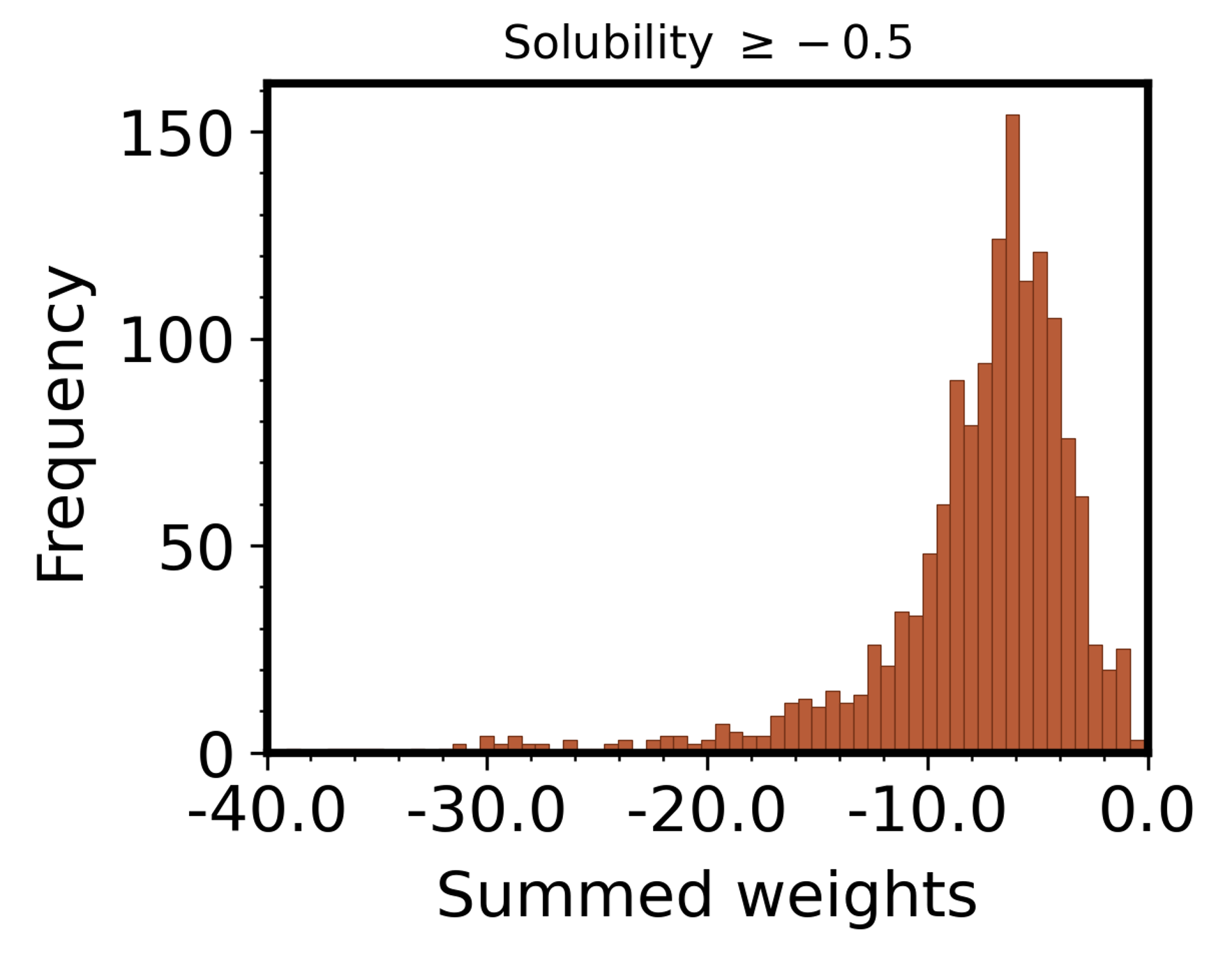}\\[-2pt]
\textnormal{(d)}
\end{minipage}
\caption{Histograms for solubility and corresponding summed embedding {summaries}
for molecules with solubility range (a,c) $\leq -5.0$ and (b,d) $\geq -0.5$,
representing the {least-soluble and most-soluble operational groups} in AqSolDB,
respectively.}
\label{fig:sol_weights}
\end{figure*}

Highly insoluble compounds are defined by $\log S \leq -5.0$ and highly soluble compounds by $\log S \geq -0.5$; these thresholds isolate the two ends of the solubility distribution. As shown in Figure~\ref{fig:sol_weights}, summed embedding {summaries} show a shift across
 two extremes of solubility. For highly soluble compounds, the distribution is concentrated in a relatively narrow range, approximately between $-10.0$ and $0.0$, with a peak near $-6.5$. Highly insoluble compounds show a much broader distribution, approximately ranging between $-35.0$ and$-7.0$, centered at
$-15.0$.

This result is an aggregate of the hidden node embeddings,
$\sum_{v}\sum_{k} h^{(L)}_{v,k}$, rather than the optimized structural output
$f(G_i)$.

\subsubsection{{GNNExplainer highlights functional groups emphasized by the structural branch}}

\begin{figure*}[htbp]
\centering
\begin{minipage}{0.47\textwidth}
\centering
\includegraphics[width=\linewidth]{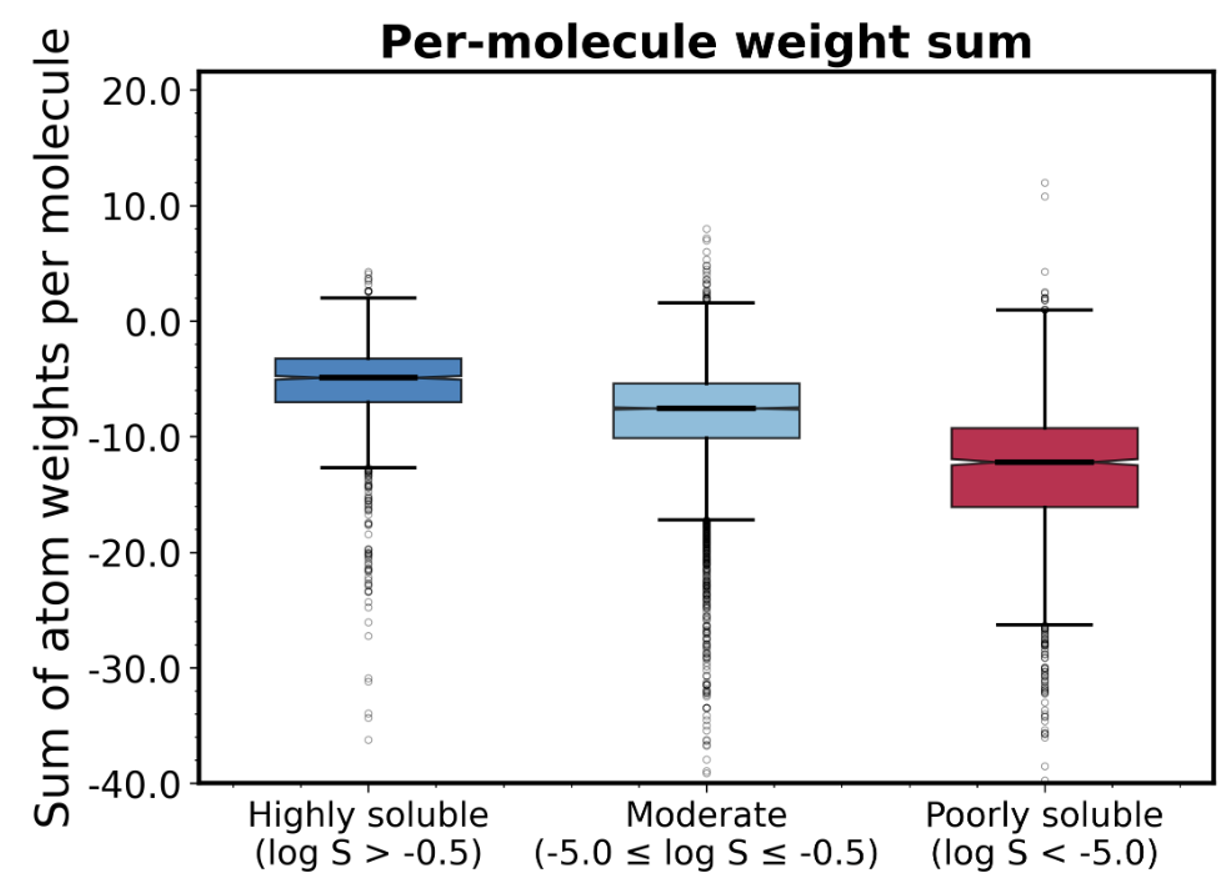}\\[-2pt]
\textnormal{(a)}
\end{minipage}
\hfill
\begin{minipage}{0.47\textwidth}
\centering
\includegraphics[width=\linewidth]{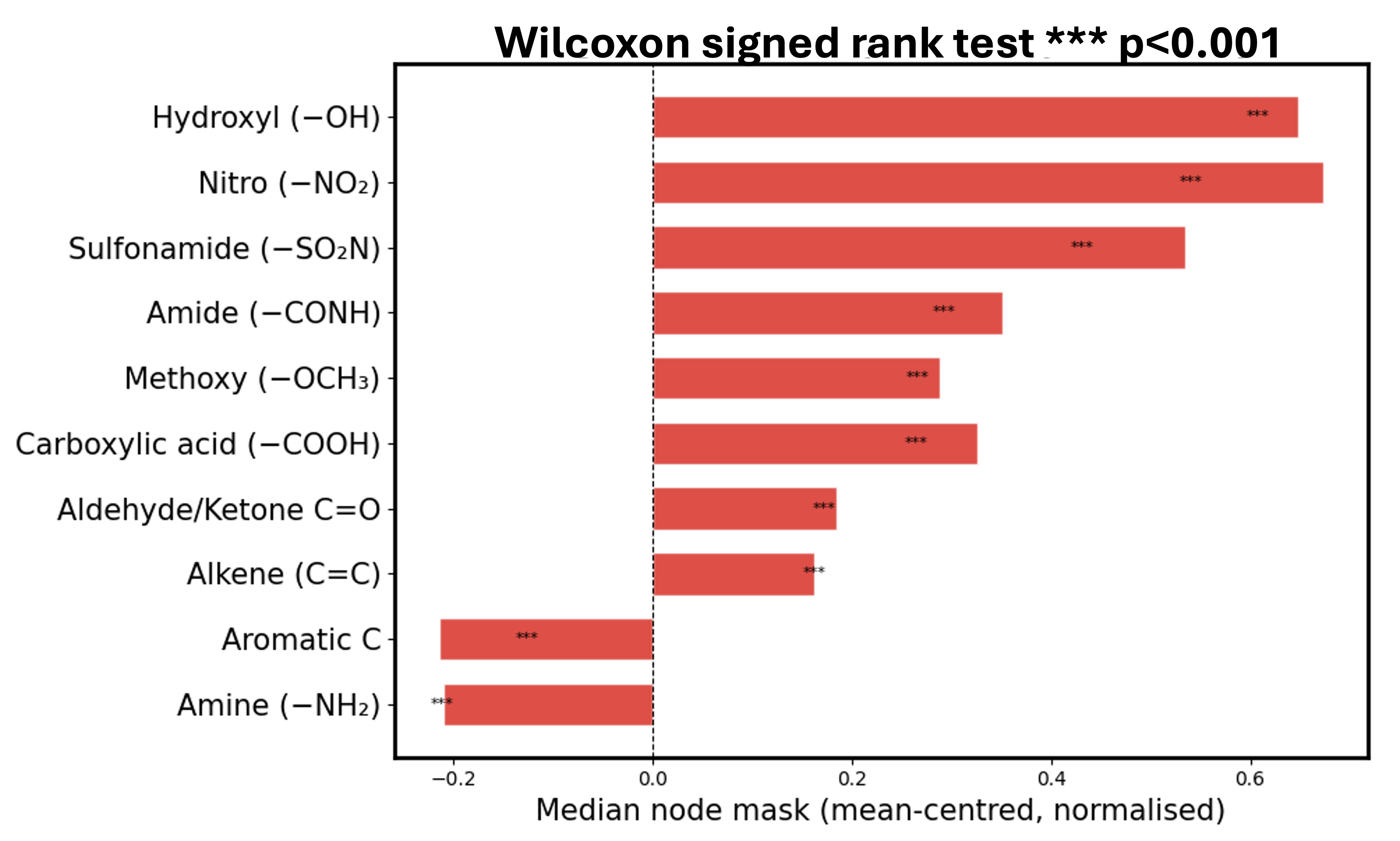}\\[-2pt]
\textnormal{(b)}
\end{minipage}
\caption{Panel (a) shows box plots of the molecule-wise embedding sums for highly soluble, moderately soluble, and poorly soluble compounds. Panel (b) presents the mean-centered median node masks as a function of functional-group fractions, with statistically significant differences denoted by asterisks (*). All embedding weights and node masks were obtained using the MPNN+MLP model.}
\label{fig:functional_group_explainer}
\end{figure*}

Finally, we apply GNNExplainer to the fitted MPNN model to identify atoms that are most important for preserving each molecule's prediction. For each molecule, the atom-level masks are mean-centered and normalized to the interval $[-1,1]$, so that each atom's value is expressed relative to the other atoms in the same molecule. {Consequently, a negative value means below the molecule-specific average mask; it does not imply a negative effect on predicted solubility.} We examine ten functional groups and substructural motifs: hydroxyl
(\ce{-OH}), nitro (\ce{-NO2}), sulfonamide (\ce{-SO2NH}), amide (\ce{-CONH}),
methoxy (\ce{-OCH3}), carboxylic acid (\ce{-COOH}), aldehyde/ketone
(\ce{C=O}), alkene (\ce{C=C}), aromatic carbon, and amine (\ce{-NH2}). These
motifs span both directions of the solubility response. The polar,
hydrogen-bonding groups---hydroxyl, carboxylic acid, amide, amine, nitro, and
sulfonamide---act as hydrogen-bond donors and/or acceptors and tend to promote
aqueous solubility, whereas the hydrophobic motifs---aromatic carbon and
alkene---tend to reduce it; the carbonyl and methoxy groups occupy an
intermediate, predominantly acceptor-like regime. This set is chemically
interpretable and balanced across the solubility response, yet small enough to
support per-group comparison, providing a basis against
which to test whether the learned structural representation $f(G_i)$ attends to
the substructures that established structure--property relationships associate
with solubility. For each functional group, we summarize every molecule containing that group by two quantities: the mean mask value of the atoms belonging to the group and the mean mask value of the remaining atoms in that same molecule. This yields one paired observation per molecule and treats the molecule, rather than the individual atom, as the unit of analysis, avoiding the artificially inflated significance that arises when correlated atoms within a molecule are treated as independent. As these per-molecule values may not be normally distributed and may contain outliers, we compare the in-group and out-of-group means using the Wilcoxon signed-rank test \cite{bauer1972constructing}, a nonparametric paired comparison of within-molecule mask-location differences. Using each molecule as its own control accounts for molecule-specific mask location. We apply this test to the ten groups listed above and report the results in Figure~\ref{fig:functional_group_explainer}(b).

In Figure~\ref{fig:functional_group_explainer}(b), the GNNExplainer analysis shows systematic differences in the importance assigned to chemically meaningful motifs. Hydroxyl, Nitro, Sulfonamide, Amide, Carboxylic acid, Aldehyde/ketone, Methoxy, and Alkene groups have positive median node-mask values, indicating that atoms in these groups are emphasized by the fitted MPNN relative to other atoms. Hydroxyl and Nitro groups show the largest positive median masks, suggesting that these motifs received higher relative within-molecule masks when forming its learned representation. In contrast, aromatic carbon and amine groups have negative median node-mask values, indicating that these atoms are relatively deemphasized by the fitted model.

These results, however, should be interpreted as local model explanations aggregated across the dataset, rather than as independent chemical effects of each functional group. Many molecules contain multiple functional groups, and the explanation masks are estimated separately for each molecule. Nevertheless, the enrichment patterns provide a chemically interpretable view of what the structural branch uses: atoms belonging to polar or strongly functionalized groups tend to receive higher importance, while some common backbone or aromatic features receive lower relative importance.

The descriptor and functional-group results should also be interpreted jointly. For example, a positive graph-branch mask for hydroxyl atoms does not by itself imply high predicted solubility: a molecule containing several hydroxyl groups may still receive an unfavorable descriptor-branch component when it also has high MolLogP, large molecular size, or extensive hydrophobic surface area. Conversely, a favorable global descriptor profile may be offset by graph motifs associated with an unfavorable structural context. As the final prediction combines both branch outputs, the net prediction reflects this balance rather than any single descriptor or functional group.

\subsubsection{Matched molecular-pair analysis}
\label{sec:mmp_analysis}

The MMP analysis uses the 8,984 molecules for which aligned descriptor- and
graph-branch outputs are available and yields 24,650 directed
pair--transformation records. Using the prespecified activity thresholds, 3,046
records (12.4\%) are descriptor-dominated, 1,101 (4.5\%) are
graph-dominated, and 10,438 (42.3\%) have appreciable contributions from both
branches. The remaining 10,065 records (40.8\%) do not satisfy these
threshold-based definitions. The category in which both branches are active
indicates that both additive components change appreciably; it does not imply
the presence of a multiplicative interaction. 

Note that the activity thresholds (less than 0.15 for an inactive branch and
greater than 0.40 for an active branch, in absolute mean change) leave an
explicit ambiguous band, so that borderline pairs are not forced into a
category. Consistent with standard MMP practice \citep{leach2006,griffen2011},
where effects are weighed against experimental uncertainty rather than a fixed
magnitude \citep{kramer2014}, these categories are intended as a conservative
descriptive summary rather than precise population estimates. Across all 24,650 pair--transformation records, the total fitted change
\(\Delta g+\Delta f\) was positively but moderately associated with the
measured change in solubility.

\begin{figure*}[t]
    \centering

    \begin{minipage}[t]{0.7\textwidth}
        \centering
        \includegraphics[width=\linewidth]{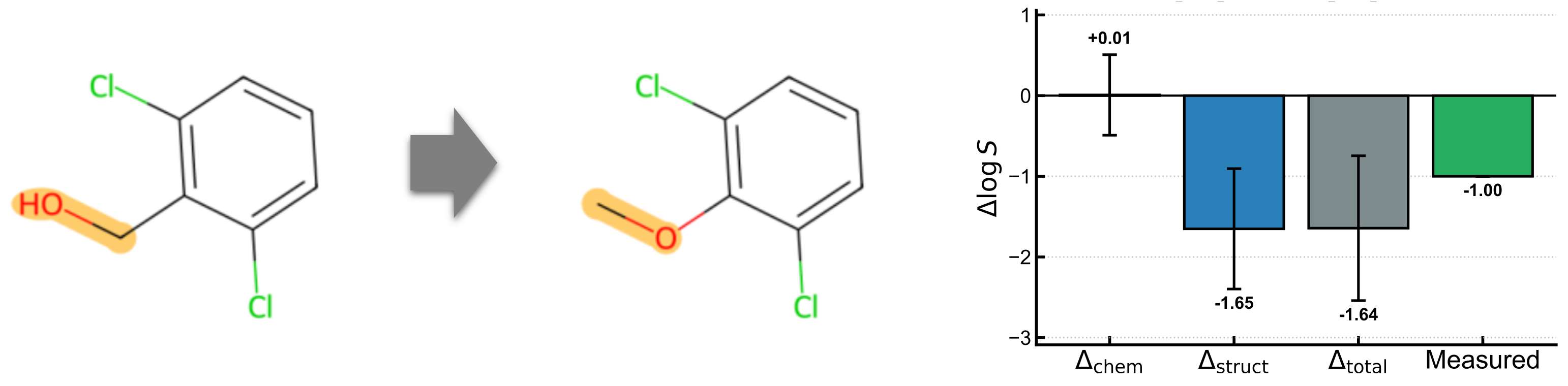}\\[-2pt]
        \textnormal{(a)}
    \end{minipage}
    \hfill
    \begin{minipage}[t]{0.7\textwidth}
        \centering
        \includegraphics[width=\linewidth]{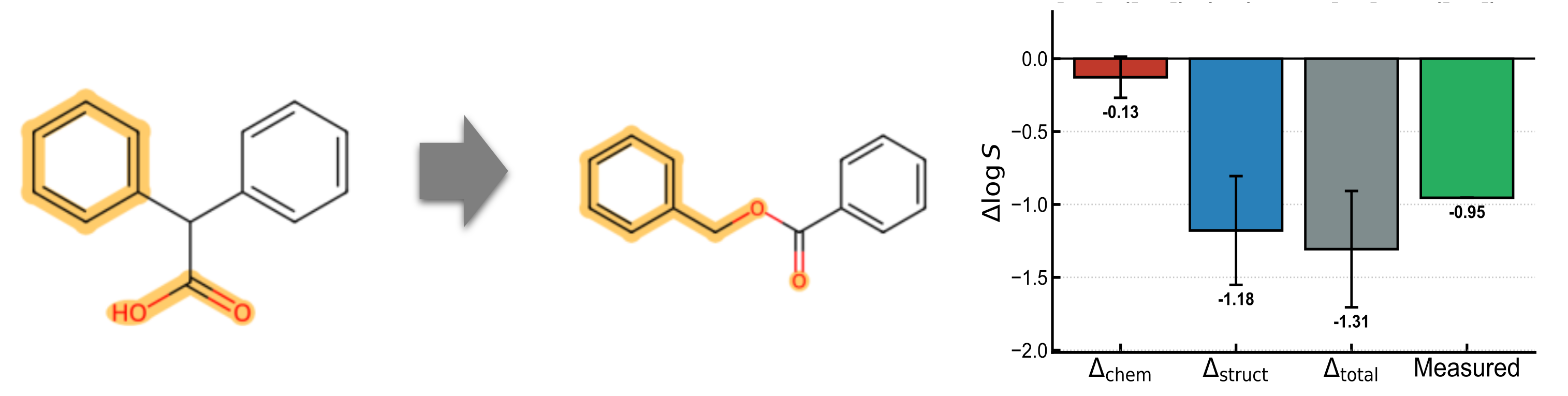}\\[-2pt]
        \textnormal{(b)}
    \end{minipage}
    \hfill
    \begin{minipage}[t]{0.7\textwidth}
        \centering
        \includegraphics[width=\linewidth]{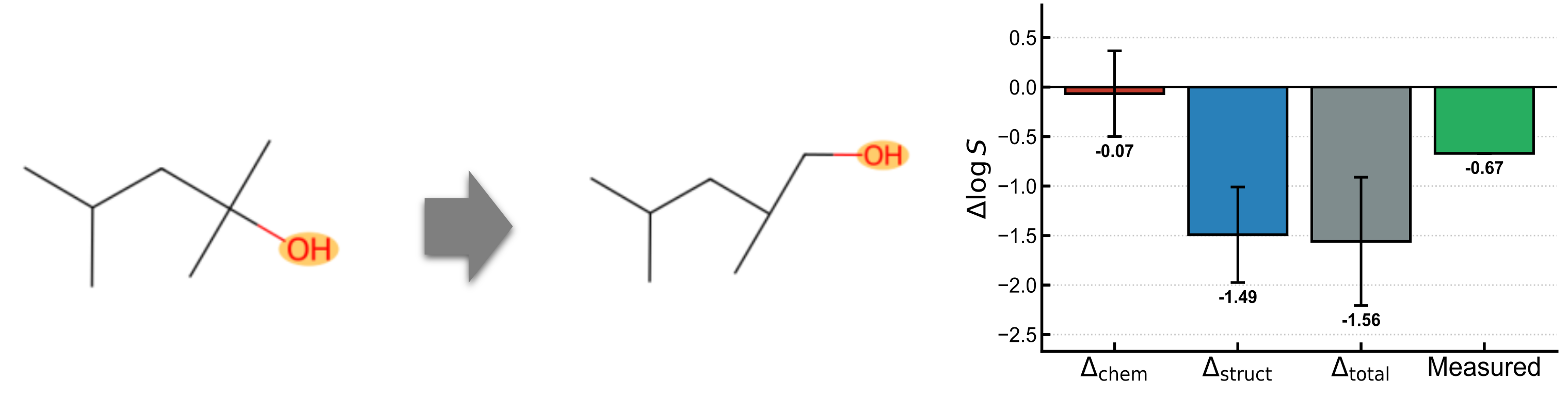}\\[-2pt]
        \textnormal{(c)}
    \end{minipage}

    \par\medskip

    \begin{minipage}[t]{0.7\textwidth}
        \centering
        \includegraphics[width=\linewidth]{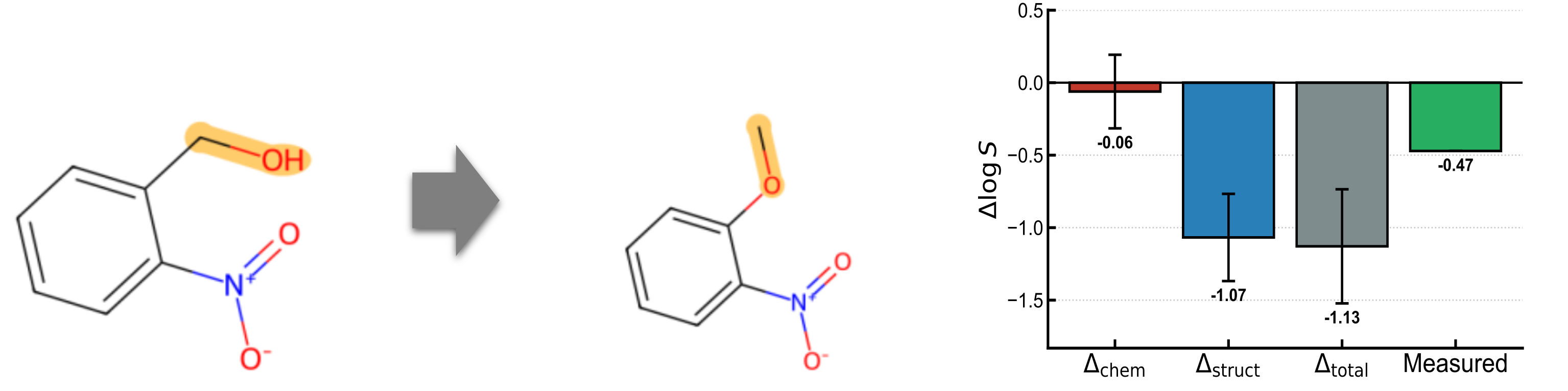}\\[-2pt]
        \textnormal{(d)}
    \end{minipage}
    \hfill
    \begin{minipage}[t]{0.7\textwidth}
        \centering
        \includegraphics[width=\linewidth]{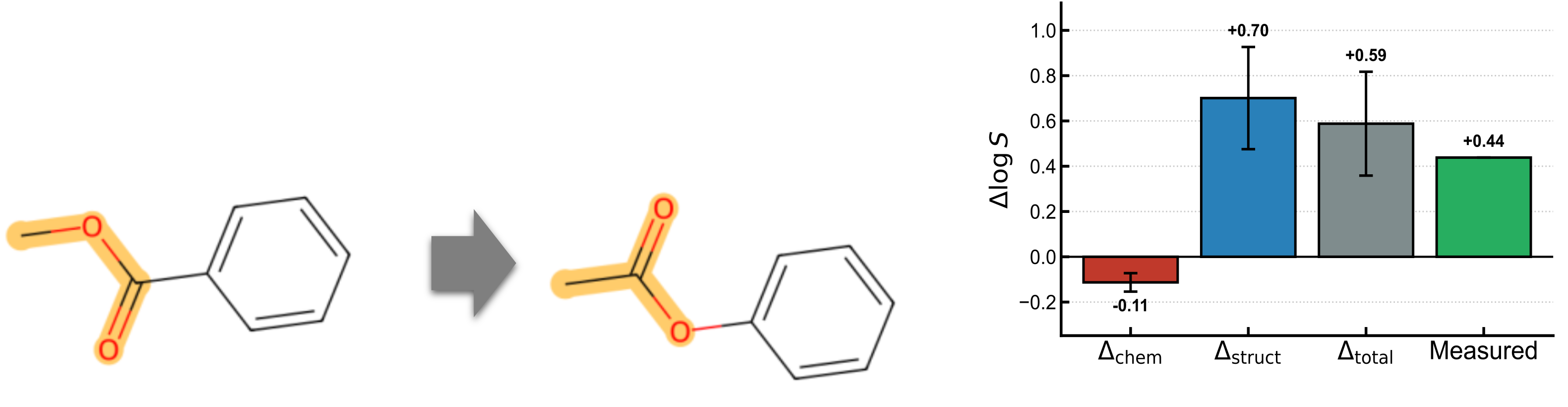}\\[-2pt]
        \textnormal{(e)}
    \end{minipage}

    \caption{Matched molecular-pair decompositions for five distinct
    same-formula (constitutional-isomer) pairs. In each panel, the two isomers
    are shown as a left-to-right transformation (arrow), with the differing
    substructure shaded in amber (the complement of their maximum common
    substructure). The bars show the mean change assigned to the
    physicochemical-descriptor branch ($\Delta g$, MLP), the molecular-graph
    branch ($\Delta f$, MPNN), their sum
    ($\Delta\widehat{y}=\Delta g+\Delta f$), and the measured change in
    $\log S$. Error bars denote one standard deviation across the ten jointly
    trained MPNN+MLP models without the interaction term.}
    \label{fig:mmp_examples}
\end{figure*}

The five constitutional-isomer pairs in
Figure~\ref{fig:mmp_examples} provide molecule-level illustrations of the
branch decomposition. For example, converting the benzylic hydroxymethyl group
of a dichlorobenzyl alcohol to a methoxy group
(Figure~\ref{fig:mmp_examples}(a);
\texttt{OCc1c(Cl)cccc1Cl}$\rightarrow$\texttt{COc1c(Cl)cccc1Cl})
produces a negligible descriptor-branch change
($\Delta g=+0.01$) but a substantially larger graph-branch change
($\Delta f=-1.65$). The resulting fitted change is
$\Delta\widehat{y}=-1.64$, compared with a measured change of $-1.00$.

Similar graph-dominated decompositions are observed for the conversion of
diphenylacetic acid to benzyl benzoate
(Figure~\ref{fig:mmp_examples}(b):
$\Delta g=-0.13$, $\Delta f=-1.18$,
$\Delta\widehat{y}=-1.31$, measured change $=-0.95$), the
tertiary-to-primary alcohol transformation
(Figure~\ref{fig:mmp_examples}(c):
$\Delta g=-0.07$, $\Delta f=-1.49$,
$\Delta\widehat{y}=-1.56$, measured change $=-0.67$), and the conversion
of a nitrobenzyl alcohol to a nitroanisole
(Figure~\ref{fig:mmp_examples}(d):
$\Delta g=-0.06$, $\Delta f=-1.07$,
$\Delta\widehat{y}=-1.13$, measured change $=-0.47$). In contrast, the
conversion of methyl benzoate to phenyl acetate
(Figure~\ref{fig:mmp_examples}(e)) produces a small negative
descriptor-branch change ($\Delta g=-0.11$) and a larger positive graph-branch
change ($\Delta f=+0.70$), yielding $\Delta\widehat{y}=+0.59$, compared with a measured change of $+0.44$. All changes are reported in $\log S$ units. Although each pair has the same molecular formula, the two molecules differ in atomic connectivity. The descriptor-branch change is small in all five examples, whereas the graph branch accounts for most of the fitted difference. The fifth pair additionally shows that the two branch contributions can have opposite signs. These examples therefore complement the population-level MBLP and GNNExplainer analyses by showing how the jointly trained branches allocate fitted solubility differences between closely related molecules.

\FloatBarrier

\section{Conclusion}

We developed an additive deep-learning framework for aqueous solubility prediction with distinct descriptor and graph input branches that are trained jointly. {In contrast to a single-input black-box approach, the architecture uses a descriptor-based MLP branch to encode physicochemical properties alongside a graph neural-network branch that encodes molecular topology.} This design provides a direct way to examine model-assigned branch contributions while retaining the flexibility of nonlinear prediction. Across AqSolDB and BigSolDB2, the simpler additive models without the multiplicative interaction term perform comparably to or better than their interaction counterparts, indicating that the two separate branches already capture most of the useful predictive information in the data.

{The framework provides directly analyzable branch outputs while maintaining competitive predictive performance. On BigSolDB2, transfer learning from AqSolDB lowered mean test MAE and reduced run-to-run variability, providing evidence of cross-dataset transfer in this dataset pair.} In our results, the non-interaction models outperformed the interaction models. The interaction term may still be useful in settings with stronger complementary signal, but its scale sensitivity and additional variance should be assessed through branch standardization or regularization in future work.

Post hoc analyses further show that the learned chemical branch aligns with
physicochemical descriptors, while the graph branch reflects interpretable
molecular topology. {Molecule-level embedding summaries} vary systematically
across solubility classes, and GNNExplainer masks {highlight} functional groups
and atom-level motifs emphasized by the fitted structural branch. The matched
molecular-pair analysis provides a complementary molecule-level interpretation:
among five distinct same-formula examples, the graph branch distinguished
changes in atomic connectivity that produced little change in the global
descriptor contribution, while the additive decomposition also revealed
whether the two branch contributions reinforced or opposed one another.
Together, these findings support the central premise of the paper: aqueous
solubility models can be made more transparent by retaining distinct descriptor
and graph pathways and examining their jointly learned outputs, while
recognizing that these outputs are neither independent nor uniquely identified
physical effects.

These findings should be interpreted within the scope of the datasets and evaluation design used here. AqSolDB and BigSolDB2 combine measurements collected under varying experimental conditions, and transfer learning was evaluated for one source–target dataset pair using scaffold-disjoint splits. In addition, because the descriptor and graph inputs are derived from the same molecules, the branch outputs quantify how the fitted model allocates information between two representations rather than uniquely defined physical effects. Evaluation on additional curated datasets will help determine how broadly these predictive and interpretive patterns extend.

Nevertheless, the proposed analysis pipeline naturally extends to several directions. The same additive framework can be implemented with more expressive graph encoders, alternative pooling mechanisms, or solvent-aware molecular representations, potentially improving predictive accuracy while retaining the branch-wise interpretation. In addition, replacing the current RDKit descriptor set with quantum-mechanical or solute--solvent interaction aware descriptors may provide a more physically grounded chemical branch. {Although physics-driven machine-learning approaches have recently shown good predictive performance supporting inference based on thermodynamics of solvation \cite{fowles2025physics,al2025accurately,amiri2026physics}, extending them to aqueous solubility prediction is not straightforward. Density functional theory (DFT) calculations can be sensitive to the choice of exchange-correlation functional, dispersion correction, solvent model, temperature, protonation state, tautomeric and stereochemical forms. These dependencies can make energy- or thermodynamics-based learning targets difficult to define consistently across large heterogeneous databases \cite{bhattacharya2024linking}. In addition, many solutes of practical interest, including zwitterions, salts, coordination complexes, and aggregating surfactants, are not well described by a neutral single-molecule representation. Thus, physics-driven extensions are an important future direction and will be explored in future work, but they require chemically consistent datasets and solution-state molecular representations.}

\newpage
\appendix

\section{Supplementary exploratory analysis}

\begin{figure*}[htbp]
    \centering
    \begin{minipage}{0.42\textwidth}
        \centering
        \includegraphics[width=\linewidth]{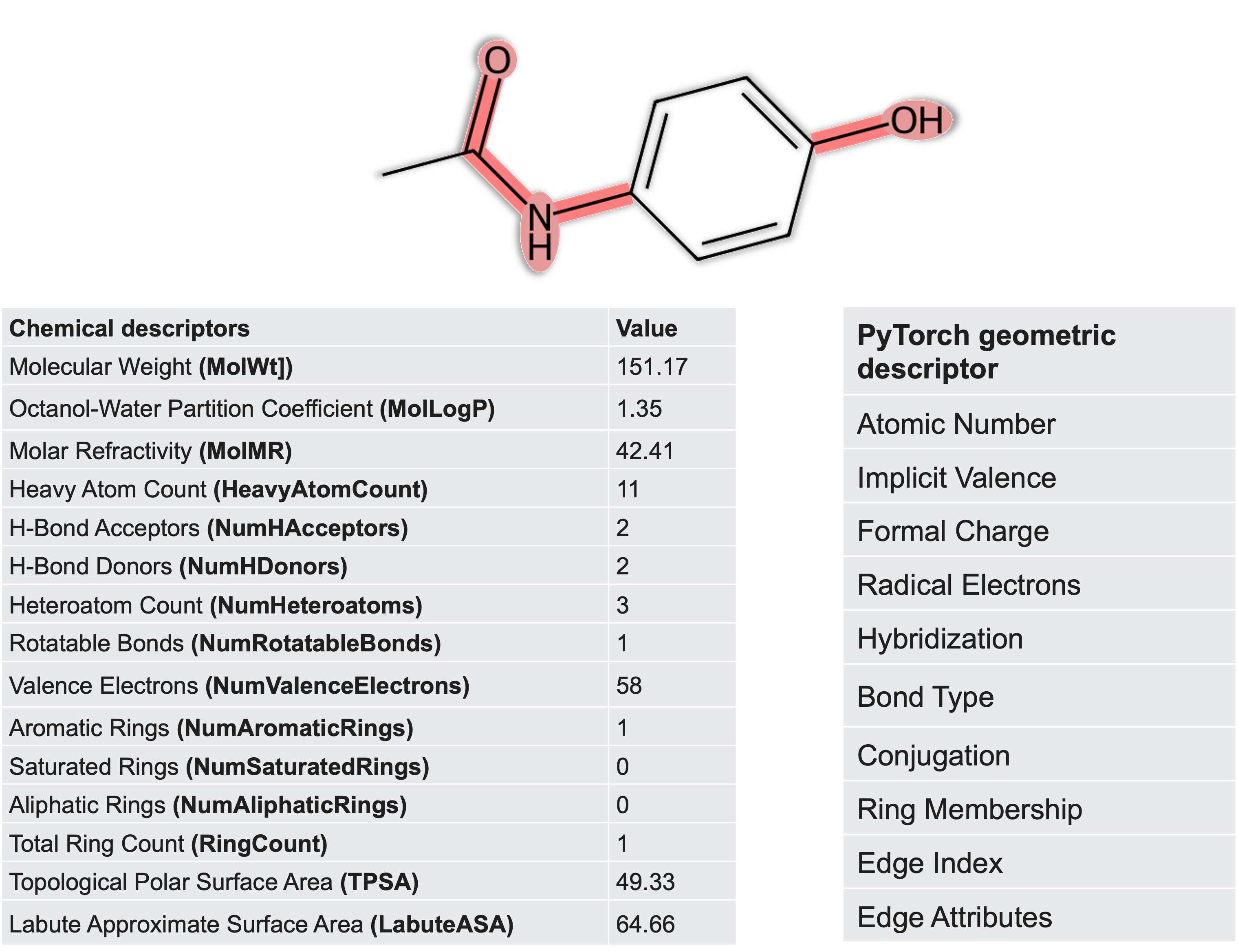}\\[-2pt]
        \textnormal{(a)}
    \end{minipage}
    \hfill
    \begin{minipage}{0.54\textwidth}
        \centering
        \includegraphics[width=\linewidth]{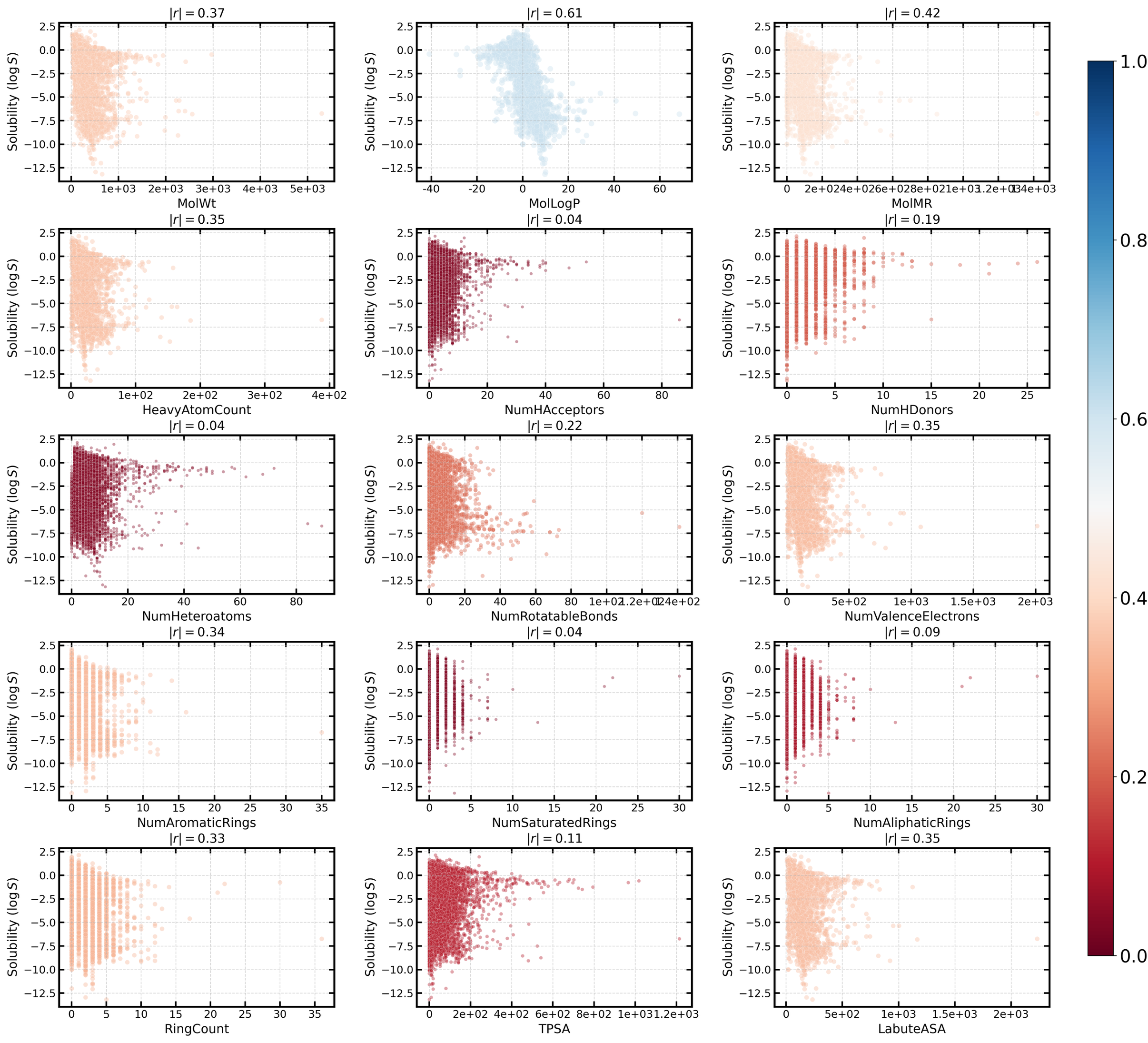}\\[-2pt]
        \textnormal{(b)}
    \end{minipage}
    \caption{(a) Schematic overview and representative example of the physicochemical descriptors and PyTorch Geometric graph features used in the proposed model. (b) Exploratory analysis: scatter plots of aqueous solubility (LogS) versus individual physicochemical descriptors for AqSolDB. The color bar and each subplot title report the magnitude of the corresponding Pearson correlation between the descriptor and LogS.}
    \label{fig:expl}
\end{figure*}

\section*{Supplementary information}
The supplementary exploratory analysis and figure are provided in Appendix~A.

\section*{Declarations}

\subsection*{Funding}
No funding was received for this work.

\subsection*{Competing interests}
The authors declare no competing interests.

\subsection*{Data and software availability}
Code for data construction, model training, transfer-learning analysis, and scaffold-split indices is
publicly available at
\url{https://github.com/spriti523/MLP-GNN-for-modeling-Aqueous-Solubility}.
The original AqSolDB dataset is available through Harvard Dataverse at
\url{https://doi.org/10.7910/DVN/OVHAW8}. The original BigSolDB~2.0 dataset is
available through Zenodo at
\url{https://doi.org/10.5281/zenodo.15094979}

\subsection*{Author contributions statement}

S.B : Conceptualization, methodology, code implementation, data curation, writing original draft, review and editing.\\
\noindent
A.R. : Methodology, writing original draft, review and editing.

\bibliographystyle{plainnat}
\bibliography{bib}

\end{document}